\documentclass[a4paper, 11pt]{article}
\usepackage[american]{babel}

\usepackage{natbib} 
\usepackage{mathtools} 
\usepackage{booktabs} 
\usepackage{tikz} 
\usepackage{enumitem}
\usepackage{algorithmic}
\usepackage{enumitem}
\usepackage[ruled]{algorithm2e}

\title{Expectation consistency for calibration of neural networks}

\usepackage[mono=false]{libertine}
\usepackage[top=2.5cm, bottom=2.5cm, left=2cm, right=2cm]{geometry}
\usepackage{amsmath, amssymb, amsthm, xfrac, mathtools, amsfonts}
\usepackage{hyperref}
\usepackage{pgfplots}
\pgfplotsset{compat=newest}
\pgfplotsset{scaled y ticks=false}
\usepgfplotslibrary{groupplots}
\usepgfplotslibrary{dateplot}
\usepackage{bm}
\usepackage{amsmath, mathtools}

\renewcommand{\vec}[1]{\bm{#1}}

\newcommand{\dd}{{\rm d}}

\newcommand{\erm}{{\rm erm}}
\newcommand{\logit}{{\rm logit}}
\newcommand{\affine}{{\rm affine}}
\newcommand{\constant}{{\rm constant}}

\newcommand{\lambdaerror}{\lambda_{\rm error}}
\newcommand{\lambdaloss}{\lambda_{\rm loss}}

\newcommand{\wstar}{\bm{w}_*}
\newcommand{\werm}{\bm{w}_{\erm}}
\newcommand{\ferm}{\hat{f}_{\erm}}
\newcommand{\fstar}{f_*}

\newtheorem{theorem}{Theorem}
\newtheorem{proposition}{Proposition}
\hypersetup{%
    colorlinks, breaklinks=true, urlcolor=orange, linkcolor=blue, citecolor=blue
}

\usepackage{authblk}

\author[1]{Lucas Clart\'e}
\author[2]{Bruno Loureiro}
\author[3]{\\Florent Krzakala}
\author[1]{Lenka Zdeborov\'a}

\affil[1]{
\'Ecole Polytechnique F\'ed\'erale de Lausanne (EPFL)\\
Statistical Physics of Computation lab.\\
CH-1015 Lausanne, Switzerland
}
\affil[2]{
D\'epartement d'Informatique, \'Ecole Normale Sup\'erieure - PSL \& CNRS, 45 rue d’Ulm, F-75230 Paris cedex 05, France
}
\affil[3]{
\'Ecole Polytechnique F\'ed\'erale de Lausanne (EPFL)\\
Information, Learning and Physics lab.\\
CH-1015 Lausanne, Switzerland
}

\begin{document}
\maketitle

\begin{abstract}
Despite their incredible performance, it is well reported that deep neural networks tend to be overoptimistic about their prediction confidence. Finding effective and efficient calibration methods for neural networks is therefore an important endeavour towards better uncertainty quantification in deep learning. In this manuscript, we introduce a novel calibration technique named \emph{expectation consistency} (EC), consisting of a post-training rescaling of the last layer weights by enforcing that the average validation confidence coincides with the average proportion of correct labels. First, we show that the EC method achieves similar calibration performance to temperature scaling (TS) across different neural network architectures and data sets, all while requiring similar validation samples and computational resources. However, we argue that EC provides a principled method grounded on a Bayesian optimality principle known as the \emph{Nishimori identity}. Next, we provide an asymptotic characterization of both TS and EC in a synthetic setting and show that their performance crucially depends on the target function. In particular, we discuss examples where EC significantly outperforms TS.
\end{abstract}

\section{Introduction}
As deep learning models become more widely employed in all aspects of human society, there is an increasing necessity to develop reliable methods to properly assess the trustworthiness of their predictions. Indeed, different uncertainty quantification procedures have been proposed to measure the confidence associated with trained neural network predictions \citep{abdar_review_2021, Gawlikowski2021}. Despite their popularity in practice, it is well known that some of these metrics, such as interpreting the last-layer softmax scores as confidence scores, lead to an overestimation of the true class probability \citep{guo_calibration_2017}. As a consequence, various methods have been proposed to calibrate neural networks \citep{gal_dropout_2016, guo_calibration_2017, maddox_simple_2019, minderer_revisiting_2021}.

In this work, we propose a novel method for the post-training calibration of neural networks named \emph{expectation consistency} (EC). It consists of fixing the scale of the last-layer weights by enforcing the average confidence to coincide with the average classification accuracy on the validation set. This procedure is inspired by optimality conditions steaming from the Bayesian inference literature. Therefore, it provides a mathematically principled alternative to similar calibration techniques such as temperature scaling, besides being simple to implement and computationally efficient. Our goal in this work is to introduce the expectation consistency calibration method, illustrate its performance across different deep learning tasks and provide theoretical guarantees in a controlled setting. More specifically, our \textbf{main contributions} are:
\begin{itemize}[wide = 1pt,noitemsep,topsep=0pt]
    \item We introduce a novel method, \textit{Expectation Consistency} (EC) to calibrate the post-training predictions of neural networks. The method is based on rescaling the last-layer weights so that the average confidence matches the average accuracy on the validation set. We provide a Bayesian inference perspective on expectation consistency that grounds it mathematically.
    \item While calibration methods abound in the uncertainty quantification literature, we compare EC to a close and widely employed method in the deep learning practice: \textit{temperature scaling} (TS). Our experiments with different network architectures and real data sets show that the two methods yield very similar results in practice. 
    \item We provide a theoretical analysis of EC in a high-dimensional logistic regression exhibiting  overconfidence issues akin to deep neural networks. We show that in this setting EC consistently outperforms temperature scaling in different uncertainty metrics. The theoretical analysis also elucidates the origin of the similarities between the two methods.
\end{itemize}
{\color{black} The manuscript is structured as follows : after a review of the literature and an exposition of the EC method (Section 3), we compare the performance of EC with TS on real data (Section 4) and show that the two methods behave similarly. In complement to Section 4, we provide in Section 5 a theoretical analysis of both methods and describe a synthetic setting in which EC outperforms TS.}

The code used in this project is available at the repository \href{https://github.com/SPOC-group/expectation-consistency}{https://github.com/SPOC-group/expectation-consistency}

\begin{algorithm}
    \centering
    \begin{algorithmic}
        \STATE {\bfseries Input:} Validation set $(\Vec{x}_{i}, y_{i})_{i = 1}^{n_{val}}$, classifier $\hat{f} : \mathcal{X} \to \mathbb{R}^K$ 
        \STATE Compute the logits $\vec{z}_{i} = \hat{f}(\vec{x}_i) \in \mathbb{R}^K$ and output $\hat{y}_{i} = \arg \max_k {\Vec{z}_{i}}_k$
        \STATE Compute the accuracy on validation set $\mathcal{A}_{val} = \frac{1}{n_{val}} \sum_{i} \delta(y_{i} = \hat{y}_{i})$ 
        \STATE Determine $T_{EC}$ such that $\frac{1}{n_{val}} \sum_{i} \max_k \sigma^{(k)}(\sfrac{\Vec{z}_{i}}{T}) = \mathcal{A}_{val}$
        \STATE {\bfseries Output:} Temperature $T_{\rm EC}$, and probabilities on new samples $\max_k \sigma^{(k)}(\sfrac{\Vec{z}^{\rm new}}{T_{\rm EC}})$, 
    \end{algorithmic}
    \caption{Expectation consistency (EC)}
    \label{algo:expectation_consistency}
\end{algorithm}

\subsection{Related work}
\paragraph{Calibration of neural networks ---} The calibration of predictive models, in particular neural networks, has been extensively studied, see \cite{abdar_review_2021, Gawlikowski2021} for two reviews. In particular, modern neural network architectures have been observed to return overconfident predictions \citep{guo_calibration_2017, minderer_revisiting_2021}. While their overconfidence could be partly attributed to their over-parametrization, some theoretical works \citep{bai_dont_2021, clarte_theoretical_2022, clarte_study_2022} have shown that even simple regression models in the under-parametrized regime can exhibit overconfidence. 

There exists a range of methods that guarantee calibration asymptotically (i.e. when the number of samples is sufficiently large) without assuming anything about the data distribution, see e.g. \cite{gupta_distribution_2020}. However, for a limited number of samples, it is less clear which of the proposed methods provides the most accurate calibration. 
\paragraph{Temperature scaling ---} \cite{guo_calibration_2017} proposed \emph{Temperature Scaling} (TS), a simple post-processing method consisting of rescaling \& cross-validating the norm of the last-layer weights. Due to its simplicity and efficiency compared to other methods such as Platt scaling \citep{platt_probabilistic_2000} or histogram binning \citep{zadrozny_obtaining_2001}, TS is widely used in practice to calibrate the output of neural networks \citep{abdar_review_2021}. Moreover, \cite{clarte_study_2022} has shown that in some settings, TS is competitive with much more costly Bayesian approaches in terms of uncertainty quantification.
While \cite{gupta_distribution_2020} has shown that without any assumption on the data model, injective calibration methods such as TS cannot be calibrated in general, \cite{guo_calibration_2017} conclude that: "\textit{Temperature scaling is the simplest, fastest, and most straightforward of the methods, and surprisingly is often the most effective.}" This justifies why TS is used so widely in practice. 
\paragraph{Bayesian methods ---} Bayesian methods such as Gaussian processes allow estimating the uncertainty out of the box for a limited number of samples under (at least implicit) data distribution assumptions. When the data-generating process is known, the best way to estimate the uncertainty of a model is to use the predictive posterior. However, Bayesian inference is often intractable, and several approximate Bayesian methods have been adapted to neural networks, such as deep ensembles \citep{Lakshminarayanan_simple_2017} or weight averaging \citep{maddox_simple_2019}. On the other hand, the strength of posthoc methods like temperature scaling is that it applies directly to the unnormalized output of the network, and does not require additional training. A comparable Bayesian approach has been developed in \cite{kristiadi_being_2020}, where a Gaussian distribution is applied to the last-layer weights. Bayesian methods typically involve sampling from a high-dimensional posterior \citep{Mattei2019}, and different methods have been proposed to compute them efficiently \citep{graves_practical_2011, gal_dropout_2016, laksh_simple_2017, maddox_simple_2019}.
\paragraph{Notation ---} We denote $[n]\coloneqq \{1,\cdots, n\}$; $\vec{1}(A)$ the indicator function of the set $A$; $\mathcal{N}(\vec{x}|\mu,\Sigma)$ the multi-variate Gaussian p.d.f. with mean $\mu$ and covariance $\Sigma$. 
\section{Setting} 
\label{sec:setting}
Consider a $K$-class classification problem where a neural network classifier is trained on a data set $(\Vec{x}_{i}, y_{i})_{i\in[n]}\in\mathbb{R}^{d}\times [K]$.  Without loss of generality, for a given input $\vec{x}\in\mathbb{R}^{d}$ we can write the output of the classifier as a $K$-dimensional vector $\vec{z}(\vec{x}) = W\vec{\varphi}(\vec{x})\in\mathbb{R}^{K}$, where we have denoted the last-layer features by $\vec{\varphi}:\mathbb{R}^{d}\to\mathbb{R}^{p}$ and the read-out weights $W\in\mathbb{R}^{K\times p}$. We define the \emph{confidence} of the prediction for class $k$ as: 
\begin{equation}
\label{eq:def:confidence}
    \hat{f}(\Vec{x}, k) \coloneqq \sigma^{(k)}(\vec{z}(\vec{x})) = \frac{e^{z_k(\vec{x})}}{\sum\limits_{l\in[K]} e^{z_l(\vec{x})}}\in (0, 1)
\end{equation}
where $\sigma:\mathbb{R}^{K}\to(0,1)^{K}$ is the softmax activation function. In short, $\hat{f}(\Vec{x}, k)$ defines a probability, as estimated by the network, that $\Vec{x}$ belongs to class $k$. For a given $\vec{x}\in\mathbb{R}^{d}$, the final \textit{prediction} of the model is then given by $\hat{y}(\Vec{x}) = \arg \max_{k} \hat{f}(\Vec{x}, k)\in [K]$, and the associated prediction confidence $\hat{f}(\Vec{x}) = \max_k \hat{f}(\Vec{x}, k) = \hat{f}(\vec{x}, \hat{y}(\vec{x}))\in(0,1)$. As it is common practice, in what follows we will be mostly interested in the case in which the network is trained by minimizing the empirical risk (ERM) with the cross-entropy loss: 
\begin{align*}
    \quad \ell(\hat{f}(\Vec{x}), y) &= - \log \hat{f}(\Vec{x}, y) \\
    &= \sum_{k = 1}^K \delta(y = k) \log \sigma^{(k)}(W\vec{\phi}(\Vec{x})),
\end{align*}
although many of the concepts introduced here straightforwardly generalize to other training procedures. The quality of the training is typically assessed by the capacity of the model to generalize on unseen data. This can be quantified by the test misclassification error and the test loss: 
\begin{equation*}
    \mathcal{E}_g = \mathbb E_{\Vec{x}, y} \left[ \delta\left(\hat{y}(\Vec{x}) \neq y \right) \right], \quad \mathcal{L}_g = - \mathbb E_{\Vec{x}, y} \left[ \log \hat{f}(\Vec{x}, y) \right]
\end{equation*}
These are point performance measures. However, often we are also interested in quantifying the quality of the network prediction confidence. Different uncertainty metrics exist in the literature, but some of the most current ones are the \textit{calibration}, \textit{expected calibration error} (ECE) and \textit{Brier score} (BS), defined as: 
\begin{align}
\label{eq:def:metrics}
\begin{cases}
    \Delta_p   &= p - \mathbb{P}_{\Vec{x}, y} \left( \hat{y}(\vec{x}) = y| \hat{f}(\Vec{x}) = p \right) \\
    \text{ECE} &= \mathbb E_{\Vec{x}, y} \left( | \Delta_{\hat{f}(\Vec{x}) } |\right) \\
    BS         &= \mathbb{E}_{\Vec{x}, y} \left( \sum_{k = 1}^K (\hat{f}(\Vec{x}, k) - \delta(y = k))^2 \right) 
\end{cases}
\end{align}

Note that the Brier score is a proper loss, meaning that it is minimized when $\hat{f}(\Vec{x}, k)$ is the true marginal distribution $\mathbb P(y = k | \vec{x})$. This is not the case of the ECE: indeed, the estimator defined as $\hat{f}(\Vec{x}, k) = \mathbb{P}(y = k)$ has $0$ ECE but does not correspond to the marginal distribution of $y$ conditioned on $\Vec{x}$ and has suboptimal test error. 
Finally, we introduce the confidence function with temperature $T>0$: 
\begin{equation}
    \hat{f}_T(\Vec{x}, k) = \sigma^{(k)}( \sfrac{W\varphi(\Vec{x})}{T}).
\end{equation}

\begin{figure*}
    \centering
    \begin{tabular}{c|c|ccc|ccc|ccc}
        \toprule
        Dataset & Model & $\mathcal{E}_g$ & $T_{TS}$ & $T_{EC}$ & $ECE$ & $ECE_{TS}$ & $ECE_{EC}$ & $BS$ & $BS_{TS}$ & $BS_{EC}$ \\
        \midrule
        SVHN & Resnet20 & 6.8 \% & 1.59 & 1.55 & 2.6 \% & 1.5 \%  & 1.3 \% & 10.5 \% & 10.4 \% & 10.4 \% \\ 
        \midrule 
        CIFAR10  & Resnet20 & 13.5 \% & 1.37 & 1.38 & 5.3 \% & 1.9 \% & 1.9 \% & 20.0 \% & 19.43\% & 19.2 \%  \\
        CIFAR10  & Resnet56 & 13.1 \% & 1.42 & 1.43 & 6.0\% & 2.5 \% & 2.4 \% & 20.2 \% & 19.3 \% & 19.3 \%  \\
        CIFAR10  & Densenet121 & 12.5 \% & 1.78 & 1.86 & 7.9 \% & 3.0 \% & 2.5 \% & 20.4 \% & 18.6 \% & 18.5 \% \\
        \midrule
        CIFAR100 & Resnet20 & 31.0 \% & 1.44 & 1.44 & 10.2 \% & 1.7 \% & 1.7 \% & 44.3 \% & 42.5 \% & 42.5 \% \\
        CIFAR100 & Resnet56 & 27.3 \% & 1.73 & 1.79 & 14 \% & 2.6 \% & 2.2 \% & 41 \% & 38.0 \% & 38.0 \%  \\ 
        CIFAR100 & VGG19 & 26.4 \% & 2.14 & 2.28 & 19.9 \% & 5.3 \% & 4.8 \% & 44.8 \% & 37.2 \% & 36.9 \% \\
        CIFAR100 & RepVGG-A2 & 22.5 \% & 1.07 & 1.16 & 5.3 \% & 4.6 \% & 4.4 \% & 32.1 \% & 31.9 \% & 32.0 \% \\
        \bottomrule
    \end{tabular}
    \caption{Comparison of expected calibration error (ECE) and Brier score (BS) of temperature scaling (TS) and expectation consistency (EC) on various models and data sets. We see very minor differences between the two calibration methods. Given how well TS works in practice we conjecture at least the same for EC.}
    \label{fig:tab_ece}
\end{figure*}

\section{Expectation consistency Calibration}
\label{sec:calibration_methods}
 The method proposed in this work acts similarly as the temperature scaling method \cite{guo_calibration_2017} discussed in the related work section, with a key difference in how the temperature parameter is chosen. The popular and widely adopted temperature scaling (TS) procedure will also serve as the main benchmark in what follows. 

\paragraph{Temperature scaling ---} Although the score-based confidence measure introduced in \eqref{eq:def:confidence} might appear natural, numerical evidence suggests that for modern neural network architectures, it tends to be overconfident \citep{guo_calibration_2017}. In other words, it overestimates the probability of class belonging. To mitigate overconfidence, \cite{guo_calibration_2017} has introduced a post-training calibration method known as \emph{temperature scaling} (TS) \citep{minderer_revisiting_2021, wang_rethinking_2021}. Temperature scaling consists of rescaling the trained network output $\vec{z}\mapsto\sfrac{\vec{z}}{T}$ by a positive constant $T>0$ (the "temperature") which is then be tuned to adjust the prediction confidence. Equivalently, TS can be seen as a re-scaling of the norm of the last-layer weights $W$. \cite{guo_calibration_2017} has found that choosing $T$ that minimizes the cross-entropy loss on the validation set $\{(\Vec{x}_i, y_i)_{i \in[n_{val}]}$: 
\begin{equation}
\label{eq:TS}
    T_{\rm TS} = \arg \min_{T>0} \left( - \sum_{i = 1}^{n_{val}} \ell(\hat{f}_T(\Vec{x}_i, y_i)) \right)
\end{equation}
results in a better calibrated rescaled predictor $\hat{f}_{T_{\rm TS}}$. To get a feeling for its effect on the confidence, it is instructive to look at the two extreme limits of TS. On one hand, if $T\ll 1$, the softmax will be dominated by the class with the largest confidence, eventually converging to a hard-thresholding $T \rightarrow 0^{+}$. This will typically lead to an overconfident predictor. On the other hand, for $T\gg 1$, the softmax will be less and less sensitive to the trained weights, converging to a uniform vector at $T\to\infty$. This will typically correspond to an underconfident predictor. Therefore, by tuning $T$, we can either make a predictor less overconfident (by lowering the temperature $T<1$) or less underconfident (by increasing the temperature $T>1$).

Temperature scaling is a specific instance of matrix/vector scaling, where the logits $z_i$ are multiplied by a matrix/vector before the softmax. Despite being more general, matrix and vector scaling have been observed in \cite{guo_calibration_2017} to perform worse than TS. Different variants of TS have been developed. Similarly to vector scaling, class-based temperature scaling \citep{frenkel_network_2021} computes one temperature per class and finds the best temperature by minimizing the validation ECE instead of the validation loss. While TS can be naturally applied to the last-layer output of neural networks, \cite{kull_beyond_2019} has extended TS to more general multi-class classification models.

\paragraph{Expectation consistency ---} In this work, we introduce a novel calibration method, which we will refer to as \emph{Expectation Consistency} (EC). As for TS, the starting point is a pre-trained confidence function $\hat{f}$ which we rescale $\hat{f}_{T}$ by introducing a temperature $T>0$. The key difference resides in the procedure we use to tune the temperature. Instead of minimizing the validation loss \eqref{eq:TS}, we search for a temperature such that the average confidence is equal to the proportion of correct labels in the test set. In mathematical terms, we define $T_{\rm EC}$ such that the following is satisfied:
\begin{equation}
\label{eq:EC}
    \frac{1}{n_{\rm val}} \sum_{i = 1}^{n_{\rm val}} \hat{f}_{T_{\rm EC}}(\Vec{x}_i) = \frac{1}{n_{\rm val}} \sum_{i = 1}^{n_{\rm val}} \vec{1}(\hat{y}(\Vec{x}_i) = y_i)
\end{equation}

\begin{figure*}[t]
    \centering
    \def\figwidth{0.3\columnwidth}
    \def\figheight{0.3\columnwidth}

\begin{tikzpicture}
\tikzstyle{every node}=[font=\tiny]

\definecolor{color0}{rgb}{0.12156862745098,0.466666666666667,0.705882352941177}
\definecolor{color1}{rgb}{1,0.647058823529412,0}

\begin{axis}[
height=\figheight,
legend cell align={left},
legend style={fill opacity=0.8, draw opacity=1, text opacity=1, draw=white!80!black},
tick align=outside,
tick pos=left,
width=\figwidth,
x grid style={white!69.0196078431373!black},
xlabel={T},
xmajorgrids,
xmin=-0.02, xmax=2.62,
xtick style={color=black},
y grid style={white!69.0196078431373!black},
ymajorgrids,
ymin=0.126989795267582, ymax=1.96145601719618,
ytick style={color=black}
]
\addplot [draw=color0, draw=none, fill=color0, forget plot, mark=square*]
table{%
x  y
1.4519901083409636 0.2103746235370636
};
\addplot [draw=color1, draw=none, fill=color1, forget plot, mark=*]
table{%
x  y
1.4598534792628888 0.9343
};
\addplot [semithick, color0]
table {%
0.1 1.8780711889267
0.182758620689655 1.03259646892548
0.26551724137931 0.716344237327576
0.348275862068966 0.552129805088043
0.431034482758621 0.452438503503799
0.513793103448276 0.386154532432556
0.596551724137931 0.339447438716888
0.679310344827586 0.305243700742722
0.762068965517241 0.279557406902313
0.844827586206896 0.259977459907532
0.927586206896552 0.244964420795441
1.01034482758621 0.233492881059647
1.09310344827586 0.224856227636337
1.17586206896552 0.218553498387337
1.25862068965517 0.214220836758614
1.34137931034483 0.211587980389595
1.42413793103448 0.210450202226639
1.50689655172414 0.210648879408836
1.58965517241379 0.212058708071709
1.67241379310345 0.214578345417976
1.7551724137931 0.218123719096184
1.83793103448276 0.222623482346535
1.92068965517241 0.228015497326851
2.00344827586207 0.234244152903557
2.08620689655172 0.241259038448334
2.16896551724138 0.249013066291809
2.25172413793103 0.257461845874786
2.33448275862069 0.26656311750412
2.41724137931035 0.276275902986526
2.5 0.286560773849487
};
\addlegendentry{Validation loss}
\addplot [semithick, color1]
table {%
0.1 0.99683278799057
0.182758620689655 0.994119822978973
0.26551724137931 0.991301476955414
0.348275862068966 0.988404870033264
0.431034482758621 0.985442876815796
0.513793103448276 0.982405006885529
0.596551724137931 0.979269206523895
0.679310344827586 0.976008176803589
0.762068965517241 0.972594499588013
0.844827586206896 0.969000220298767
0.927586206896552 0.965198874473572
1.01034482758621 0.961165249347687
1.09310344827586 0.956876277923584
1.17586206896552 0.952310144901276
1.25862068965517 0.947447776794434
1.34137931034483 0.9422727227211
1.42413793103448 0.936771333217621
1.50689655172414 0.930932760238647
1.58965517241379 0.924749910831451
1.67241379310345 0.918218851089478
1.7551724137931 0.911338865756989
1.83793103448276 0.904112875461578
1.92068965517241 0.896546959877014
2.00344827586207 0.888650417327881
2.08620689655172 0.880434811115265
2.16896551724138 0.871914923191071
2.25172413793103 0.86310738325119
2.33448275862069 0.854031085968018
2.41724137931035 0.844706356525421
2.5 0.835154592990875
};
\addlegendentry{Average confidence}
\addplot [semithick, black, dashed]
table {%
0.1 0.934293429342934
0.182758620689655 0.934293429342934
0.26551724137931 0.934293429342934
0.348275862068966 0.934293429342934
0.431034482758621 0.934293429342934
0.513793103448276 0.934293429342934
0.596551724137931 0.934293429342934
0.679310344827586 0.934293429342934
0.762068965517241 0.934293429342934
0.844827586206896 0.934293429342934
0.927586206896552 0.934293429342934
1.01034482758621 0.934293429342934
1.09310344827586 0.934293429342934
1.17586206896552 0.934293429342934
1.25862068965517 0.934293429342934
1.34137931034483 0.934293429342934
1.42413793103448 0.934293429342934
1.50689655172414 0.934293429342934
1.58965517241379 0.934293429342934
1.67241379310345 0.934293429342934
1.7551724137931 0.934293429342934
1.83793103448276 0.934293429342934
1.92068965517241 0.934293429342934
2.00344827586207 0.934293429342934
2.08620689655172 0.934293429342934
2.16896551724138 0.934293429342934
2.25172413793103 0.934293429342934
2.33448275862069 0.934293429342934
2.41724137931035 0.934293429342934
2.5 0.934293429342934
};
\addlegendentry{Validation accuracy}
\end{axis}

\end{tikzpicture} 
\begin{tikzpicture}
\tikzstyle{every node}=[font=\tiny]

\definecolor{color0}{rgb}{0.12156862745098,0.466666666666667,0.705882352941177}
\definecolor{color1}{rgb}{1,0.498039215686275,0.0549019607843137}

\begin{axis}[
height=\figheight,
tick align=outside,
tick pos=left,
width=\figwidth,
x grid style={white!69.0196078431373!black},
xlabel={T},
xmajorgrids,
xmin=-0.02, xmax=2.62,
xtick style={color=black},
y grid style={white!69.0196078431373!black},
ylabel={ECE},
ymajorgrids,
ymin=0.00215900007262826, ymax=0.115645934706181,
ytick style={color=black}
]
\addplot [draw=color0, draw=none, fill=color0, mark=square*, mark size=2pt]
table{%
x  y
1.4519901083409636 0.04489138384163379
};
\addplot [draw=color1, draw=none, fill=color1, mark=*, mark size=2pt]
table{%
x  y
1.4598534792628888 0.0444057314068079
};
\addplot [semithick, black]
table {%
0.1 0.110487437677383
0.182758620689655 0.107043664851785
0.26551724137931 0.103518988212943
0.348275862068966 0.0999590999484062
0.431034482758621 0.0963715692609548
0.513793103448276 0.092749162620306
0.596551724137931 0.0890801307320595
0.679310344827586 0.0853508675038814
0.762068965517241 0.0815465277612209
0.844827586206896 0.0776519088119268
0.927586206896552 0.0736508421987295
1.01034482758621 0.0695283965289593
1.09310344827586 0.0652693948745727
1.17586206896552 0.0608693632155657
1.25862068965517 0.0562855450451374
1.34137931034483 0.0515350782573223
1.42413793103448 0.0465971593976021
1.50689655172414 0.0414621486216783
1.58965517241379 0.0361220541983843
1.67241379310345 0.0305705555260181
1.7551724137931 0.0252937272399664
1.83793103448276 0.0209881811618805
1.92068965517241 0.0163609291434288
2.00344827586207 0.0116558670192957
2.08620689655172 0.00731749710142612
2.16896551724138 0.00936922290027142
2.25172413793103 0.014536472800374
2.33448275862069 0.0218473352134228
2.41724137931035 0.0293591247737408
2.5 0.0370628777205944
};
\end{axis}

\end{tikzpicture}
                \hspace{6mm}
                \includegraphics[width=0.25\columnwidth]{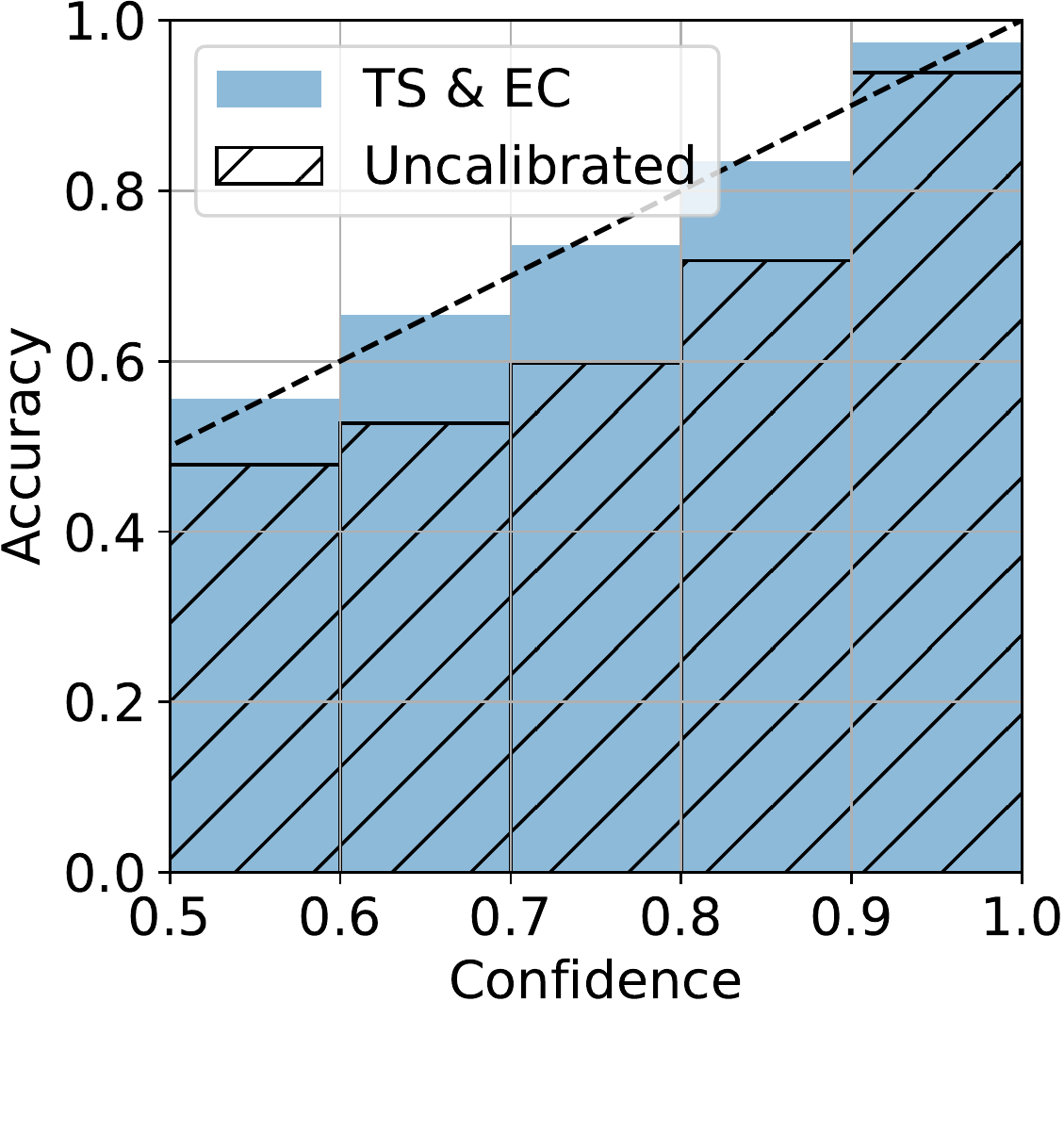}
    
    \caption{Left: The validation loss and average confidence of the model, as a function of the temperature $T$, model is DenseNet121 trained on CIFAR10. The dark dashed line is the accuracy for the validation set. Orange (respectively blue) cross corresponds to $T_{EC}$, $T_{TS}$. Middle: ECE of the model as a function of $T$, blue and orange dots respectively correspond to TS and EC. Right: Reliability diagram of Resnet20 trained on CIFAR10, before and after Temperature scaling. The reliability diagram after EC is indistinguishable from the one of TS.}

    \label{fig:real_data_curves}
\end{figure*}
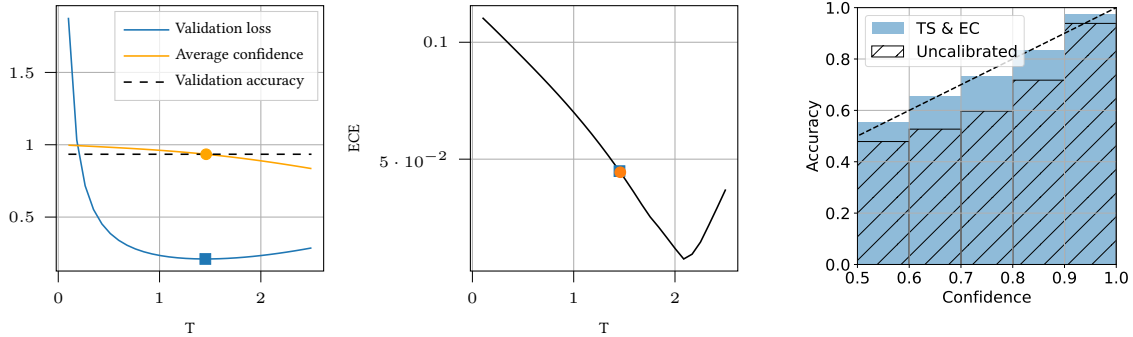

The intuition behind this choice is the following: a calibrated classifier is such that for all $p \in (0, 1), \Delta_p = 0$. This condition is not achievable by tuning the temperature parameter $T$, so a less strict condition is to enforce it in expectation $\mathbb{E}_{\vec{x}} \left[ \Delta_{ \hat{f}(\vec{x})} \right] = 0$, ensuring that the classifier is calibrated on average. This is equivalent to enforcing the average confidence to be equal to the probability of predicting the correct class on a validation set. Note that the fact that we directly compare to the confidence on the validation set is analogous to what is done in the conformal prediction \citep{papadopoulos2002inductive} methods to estimate prediction sets (as opposed to calibration that we are aiming at here). 

It is instructive to consider a Bayesian perspective on EC. For the sake of this paragraph, assume that both the training and validation data were independently drawn from a parametric probability distribution $p(\vec{x}, y|\theta)$. If we had access to the distribution of the data (but not the specific realization of the parameters $\theta$), the Bayes-optimal confidence function would be given by the expectation of $f_{\star}(\vec{x}|\theta) = p(y|\vec{x},\theta)$ with respect to the posterior distribution of the weights given the training data $p(\theta|(\vec{x}_{i},y_{i})_{i\in [n]})$. In this case, one would not even need a validation set since the expected test accuracy would be predicted by the uncertainties under the posterior. In Section \ref{sec:bayesian} we illustrate this discussion for concrete data distribution. 
This expectation consistency property of the Bayes-optimal predictor is known as the \emph{Nishimori condition} in the information theory and statistical physics literature \citep{iba1999nishimori,measson2009generalized,zdeborova2016statistical}. Therefore, from this perspective requesting condition \eqref{eq:EC} to hold can be seen as enforcing the Nishimori conditions for the rescaled confidence function.
The Nishimori conditions are also used within the expectation-maximization algorithm for learning hyperparameters \cite{dempster1977maximum}. We describe in Section \ref{sec:theory_method} how to interpret both temperature scaling and expectation consistency as learning procedures for the hyperparameter $T$. 
The main idea behind the EC method proposed here is that even in the absence of knowledge of the data-generating model, the expectation consistency \eqref{eq:EC} relation should hold for a calibrated uncertainty quantification method.


Note that $T_{EC}$ exists and is unique. Indeed, the average confidence is a decreasing function of the temperature, converging to one when $T \to 0^{+}$ and to zero when $T \to \infty$. Therefore, there is a unique $T_{EC}$ that satisfies the constraint \eqref{eq:EC}, and in practice, it can be found by bisection. We refer to Figure~\ref{fig:real_data_curves} for an illustration of the uniqueness of $T_{EC}$. Moreover, note that expectation consistency is more flexible than temperature scaling: in multi-class classification problems, we can fix the temperature so that the average confidence is equal to the top $N$ accuracy for any $N\in [K]$. In this work, we focus on the top $1$ accuracy. 


\section{Experiments on real data}
\label{sec:real_data}

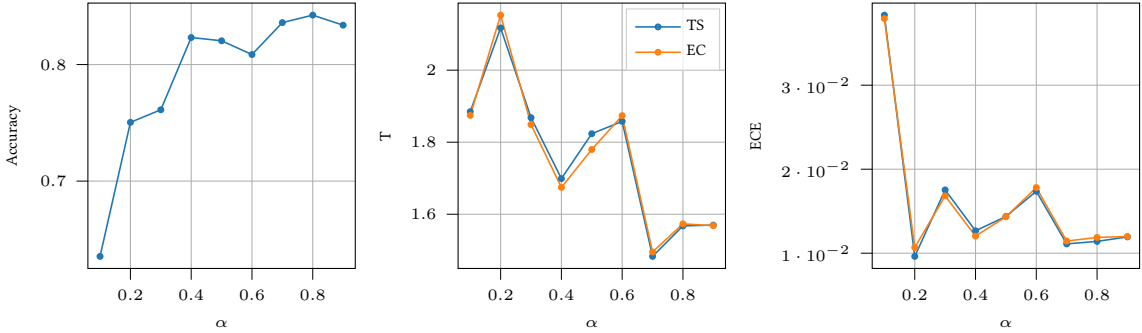
\begin{figure*}[t]
    \centering
    \def\figwidth{0.3\columnwidth} 
    \def\figheight{0.3\columnwidth}

\begin{tikzpicture}
\tikzstyle{every node}=[font=\tiny]

\definecolor{color0}{rgb}{0.12156862745098,0.466666666666667,0.705882352941177}

\begin{axis}[
height=\figheight,
tick align=outside,
tick pos=left,
width=\figwidth,
x grid style={white!69.0196078431373!black},
xmajorgrids,
xmin=0.06, xmax=0.94,
xtick style={color=black},
y grid style={white!69.0196078431373!black},
ymajorgrids,
ymin=0.62505, ymax=0.85275,
ytick style={color=black},
xlabel={$\alpha$},
ylabel={Accuracy},
]
\addplot [semithick, color0, mark=*, mark size=1, mark options={solid}]
table {%
0.1 0.6354
0.2 0.7504
0.3 0.7612
0.4 0.8232
0.5 0.8204
0.6 0.8086
0.7 0.836
0.8 0.8424
0.9 0.8338
};
\end{axis}

\end{tikzpicture}
\begin{tikzpicture}
\tikzstyle{every node}=[font=\tiny]

\definecolor{color0}{rgb}{0.12156862745098,0.466666666666667,0.705882352941177}
\definecolor{color1}{rgb}{1,0.498039215686275,0.0549019607843137}

\begin{axis}[
height=\figheight,
legend cell align={left},
legend style={fill opacity=0.8, draw opacity=1, text opacity=1, draw=white!80!black},
tick align=outside,
tick pos=left,
width=\figwidth,
x grid style={white!69.0196078431373!black},
xlabel={\(\displaystyle \alpha\)},
xmajorgrids,
xmin=0.06, xmax=0.94,
xtick style={color=black},
y grid style={white!69.0196078431373!black},
ylabel={T},
ymajorgrids,
ymin=1.44952724523415, ymax=2.18619021162545,
ytick style={color=black}
]
\addplot [semithick, color0, mark=*, mark size=1, mark options={solid}]
table {%
0.1 1.88481291932946
0.2 2.11724267706903
0.3 1.86816397818417
0.4 1.69894991197592
0.5 1.8235535146923
0.6 1.85783313425149
0.7 1.48301192552466
0.8 1.56764228777125
0.9 1.56987584777543
};
\addlegendentry{TS}
\addplot [semithick, color1, mark=*, mark size=1, mark options={solid}]
table {%
0.1 1.8744440482067
0.2 2.15270553133494
0.3 1.84908698883878
0.4 1.67449117965848
0.5 1.77987916864334
0.6 1.87372293391687
0.7 1.49526720693508
0.8 1.57330057692271
0.9 1.5683016778546
};
\addlegendentry{EC}
\end{axis}

\end{tikzpicture}
\begin{tikzpicture}
\tikzstyle{every node}=[font=\tiny]

\definecolor{color0}{rgb}{0.12156862745098,0.466666666666667,0.705882352941177}
\definecolor{color1}{rgb}{1,0.498039215686275,0.0549019607843137}

\begin{axis}[
height=\figheight,
legend cell align={left},
legend style={fill opacity=0.8, draw opacity=1, text opacity=1, draw=white!80!black},
tick align=outside,
tick pos=left,
width=\figwidth,
x grid style={white!69.0196078431373!black},
xlabel={\(\displaystyle \alpha\)},
xmajorgrids,
xmin=0.06, xmax=0.94,
xtick style={color=black},
y grid style={white!69.0196078431373!black},
ylabel={ECE},
ymajorgrids,
ymin=0.00818590496987105, ymax=0.0397754454770684,
ytick style={color=black}
]
\addplot [semithick, color0, mark=*, mark size=1, mark options={solid}]
table {%
0.1 0.0383395572721958
0.2 0.00962179317474365
0.3 0.0175336042165756
0.4 0.0126675539493561
0.5 0.0143727298617363
0.6 0.0173717992067337
0.7 0.0111104053735733
0.8 0.0114077365040779
0.9 0.0119451258897781
};
\addplot [semithick, color1, mark=*, mark size=1, mark options={solid}]
table {%
0.1 0.0379269671022892
0.2 0.0106787579894066
0.3 0.0168228935480118
0.4 0.0120353922367096
0.5 0.014335719704628
0.6 0.0178179466485977
0.7 0.0114488594293594
0.8 0.0118761380434036
0.9 0.0119916478991508
};
\end{axis}

\end{tikzpicture}
    \caption{Left: Accuracy of Resnet20 model (Left), the temperature returned by TS and EC (Middle) and ECE of the model (Right) as a function of the size of the training set $\alpha = \sfrac{n_{\rm train}}{50000}$. The model is trained with the same hyperparameters as in Figure~\ref{fig:tab_ece}. Again we see that the two methods are comparable even at largely different sample sizes.}
    \label{fig:train_ratio}.    
\end{figure*}

In this section, we present numerical experiments carried out on real data sets and compare the performance of EC and TS. As we will see, both methods yield similar calibration performances in practical scenarios.

\label{sec:experimental_setup}
\paragraph{Experimental setup --- } We consider the performance of the calibration methods from Section~\ref{sec:calibration_methods} in image classification tasks. Experiments were conducted on three popular image classification data sets: 
\begin{itemize}[wide = 1pt,noitemsep,topsep=0pt]
    \item SVHN \cite{netzer_reading_2011} is made of colored $32 \times 32$ labelled digit images. Train/validation/test set sizes are 65931/7325/26032.
    \item CIFAR10 and CIFAR100 data sets \cite{krizhevsky_learning_2009}, consisting of $32 \times 32$ colored images from 10/100 classes (dog, cat, plane, etc.), respectively. Train/validation/test sets sizes are 45000/5000/10000 images for CIFAR10, 50000/5000/5000 for CIFAR100.
\end{itemize}
We consider different neural network architectures adapted to image classification tasks: ResNets \citep{he_resnet_2016}, DenseNets \citep{huang_densely_2017}, VGG \citep{simonyan_very_2014}  and RepVGG \citep{ding_repvgg_2021}. For CIFAR100, pre-trained models available online were employed. More details on the training procedure are available in Appendix A.

\paragraph{Results ---} We refer to Table~\ref{fig:tab_ece} for a comparison of TS and EC on the various data sets and models discussed above. Curiously, we observe that both EC and TS yield very similar temperatures across the different tasks and architectures, implying a similar ECE and Brier score. In particular, note that both methods give $T>1$, consistent with the fact that the original networks were overconfident. Therefore, as expected, both methods improve the calibration of the classifiers. 

The right panel of Figure~\ref{fig:real_data_curves} shows the reliability diagram of the ResNet 20 trained on CIFAR10: we observe that before applying TS and EC, the accuracy is lower than the confidence. In other words, the model is overconfident and both TS and EC improve the calibration of the model.

Note that both methods improve the Brier score and yield very similar results. From the computational cost perspective, EC is as efficient to run as TS, and requires only a few lines of code, see the \href{https://github.com/SPOC-group/expectation-consistency/}{GitHub} repository where we provide the code to reproduce the experiments discussed here. However, we believe expectation consistency is a more principled calibration method, as it constrains the confidence of the model to correspond to the accuracy and has a natural Bayesian interpretation. Moreover, as we will discuss in Section \ref{sec:theory_method}, we can derive explicit theoretical results for EC.

Our experiments suggest that the similarity between TS and EC is independent of the accuracy of the model. Indeed, in Figure~\ref{fig:train_ratio}, we observe the accuracy and ECE of a ResNet model trained on different amounts of data. As expected, the accuracy of the model increases with the amount of training data. We observe in the middle and right panels that the temperatures and ECE obtained from both methods are extremely similar, independently of the accuracy of the model. Finally, we plot in the middle panel of Figure~\ref{fig:real_data_curves} the ECE as a function of the temperature and observe that neither $T_{TS}$ nor $T_{EC}$ is close to the minimum of ECE. However, as we have discussed in Section \ref{sec:calibration_methods}, ECE is only one uncertainty quantification metric and is not a proper loss, so we wish not to optimize the temperature for this metric in particular.

{\color{black}
\paragraph{Experiments on corrupted data ---}

In Appendix C we compare the performance of EC and TS on an image classification task where the test data is corrupted. We use the same datasets and architectures as in Section~\ref{sec:real_data}, but for some image classes, the target labels on the test data are randomly chosen. The goal of introducing a class-dependent noise is to evaluate both methods in a more realistic scenario where there is a distribution shift between the training and test data, as done in \cite{hendrycks_benchmarking_2019}. We report in Table 1 of the Appendix the performance of EC and TS in terms of ECE and Brier score. We observe that EC yields an reduction of the test ECE of 7 \% on average, showing that EC is a competitive alternative to TS in more realistic scenarios. Note that in this setting, EC and TS yield different temperatures, contrary to the results described in Table~\ref{fig:tab_ece}. 
The full experimental details are described in Appendix C.
}
\section{Theoretical analysis of the EC}
\label{sec:theory_method}
As we have seen in Section \ref{sec:real_data}, our experiments with real data and neural network models suggest that despite their different nature, EC and TS achieve a similar calibration performance across different architectures and data sets. In this section, we investigate EC and TS in specific settings where we can derive theoretical guarantees on their calibration properties.

\begin{figure*}[!ht]
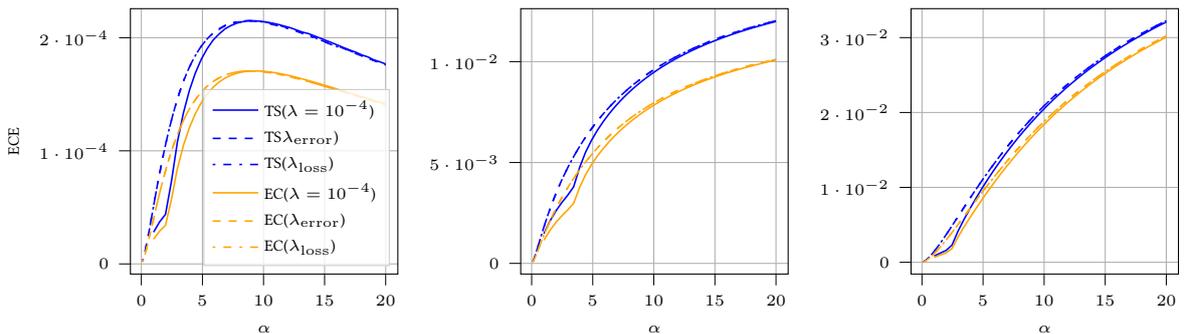

    \centering
    \def\figwidth{0.3\columnwidth} 
    \def\figheight{0.3\columnwidth}

    \input{logit_teacher/logit_teacher_ece.tex}
    \input{affine_teacher/affine_teacher_ece.tex}
    \input{constant_teacher/constant_teacher_ece.tex}
    
    \caption{ECE of regularized logistic regression with three different values of $\lambda$ ($10^{-4}, \lambda_{\rm error}, \lambda_{\rm loss}$): uncalibrated, after temperature scaling, and after expectation consistency. From left to right: $\sigma_{\star} = \sigma_{\logit}, \sigma_{\affine},\sigma_{\constant}$ respectively.}
    \label{fig:ece}
\end{figure*}

For concreteness, in the examples that follow we will focus on binary classification problems for which, without loss of generality, we can assume $y\in\{-1,+1\}$. In this encoding, the softmax function is equivalent to the logit $\sigma(t)\coloneqq (1+e^{-t})^{-1}$, and the hard-max is given by the sign function. Further, we assume that both the training $(\vec{x}_{i}, y_{i})_{i\in[n]}$ and validation set $(\vec{x}_{i}, y_{i})_{i\in[n_{val}]}$ were independently drawn from the following data generative model:
\begin{align}
\label{eq:def:data}
    f_{\star}(\vec{x}) &\coloneqq \mathbb{P}( y^{\mu} = 1 | \vec{x}^\mu) = \sigma_{\star} \left( \frac{\wstar^{\top}\vec{x}^{\mu}}{T_{\star}} \right) \notag\\
    \vec{x}^{\mu} &\sim\mathcal{N}(\vec{0},\sfrac{1}{d}\mathbf{I}_{d}), \quad 
\wstar\sim\mathcal{N}(\vec{0},\mathbf{I}_{d})
\end{align}
with $\sigma_{\star}:\mathbb{R}\to (0,1)$ an activation function and $T_{*}>0$ explicitly parametrizing the norm of the weights. 

First, in Section \ref{sec:bayesian} we provide a Bayesian interpretation of both the TS and EC methods, in an example where $T_{\rm{TS}} = T_{\rm{EC}}=T_{\star}$. Next, in Section \ref{sec:theory:misspecified} we analyze a misspecified empirical risk minimization setting where they yield different results. Finally, we discuss in Section \ref{sec:theory:outperforms} one case in which EC consistently outperforms TS. 
\subsection{Relation with Bayesian estimation} 
\label{sec:bayesian}
Consider a Bayesian inference problem: given the training data $\mathcal{D}\coloneqq \{(\vec{x}_{i}, y_{i})_{i\in[n]}\}$, what is the predictor that maximizes the accuracy? If the statistician had complete access to the data generating process \eqref{eq:def:data}, this would be given by integrating the likelihood of the data over the posterior distribution of the weights given the data:
\begin{align*}
    f_{\rm{bo}}(\vec{x}) \coloneqq \mathbb{P}(y=1|\mathcal{D},\vec{x}) = \int\dd\vec{w}~\sigma\left(\frac{\vec{w}^{\top}\vec{x}}{T_{*}}\right)p(\vec{w}|\mathcal{D}, T_{\star})
\end{align*}
where the posterior distribution is explicitly given by:
\begin{align}
\label{eq:posterior}
    p(\vec{w} | \mathcal{D}, T_{\star}) \propto \mathcal{N}(\vec{w}|0, \mathbf{I}_{d})\prod\limits_{i\in [n]}\sigma_{*}\left(y_{i}\frac{\vec{w}^{\top}\vec{x}_{i}}{T_{\star}}\right).
\end{align}
The Bayes-optimal predictor above is well calibrated \citep{clarte_theoretical_2022}, and consequently satisfies the expectation consistency condition: its average confidence equates to its accuracy. Consider now a scenario where the statistician only has {\it partial} information about the data-generating process: she knows the prior and likelihood but does not have access to the true temperature $T_{\star}$. In this case, she could still write a posterior distribution but would need to estimate the temperature $T$ from the data. This can be done by finding the $T$ that minimizes the classification error, or equivalently the generalisation loss, yet equivalently this would correspond to expectation maximization as discussed e.g. in \citep{decelle2011asymptotic,krzakala2012probabilistic}. This estimation of the temperature would lead to $T = T_*$ and  recovers the Bayes-optimal estimator $f_{\rm{bo}}$. Hence, in the \textit{well-specified} Bayesian setting, doing temperature scaling amounts to expectation consistency, providing a very natural interpretation of both the temperature scaling and expectation consistency methods in a Bayesian framework.

Note that in this paper we are concerned with frequentist estimators trained via empirical risk minimization. In that case, even in the well-specified setting, neither TS nor EC will recover the correct temperature $T_*$ in the high-dimensional limit. This impossibility to recover $T_*$ comes from the fact that we are not sampling a distribution anymore but instead consider a point estimate and do not have enough samples to be in the regime where point estimators are consistent.

\subsection{Misspecified ERM}
\label{sec:theory:misspecified}
Consider now the case in which the statistician only has access to the training data $\mathcal{D}$, with no knowledge of the underlying generative model. A popular classifier for binary classification in this case is logistic regression, for which:
\begin{align}
\label{eq:def:logit}
    \hat{f}_{\erm}(\vec{x}) = \sigma(\hat{\vec{w}}^{\top}\vec{x})
\end{align}
and the weights are obtained by minimizing the empirical risk over the training data $\hat{\vec{w}} = {\rm argmin}\hat{\mathcal{R}}_n(\vec{w})$ where:
\begin{align}
\label{eq:def:risk}
\hat{\mathcal{R}}_n(\vec{w}) &= - \sum_{i\in[n]} \log \sigma ( \vec{w}^{\top} \vec{x} ) + \sfrac{\lambda}{2} \| \vec{w} \|^2
\end{align}
and we remind that $\sigma$ is the sigmoid/logit function. In this setting, the calibration is given by $\Delta_{\ell} = \ell - \mathbb E_{\vec x} \left[ f_*(\Vec{x}) | \ferm(\vec x) = \ell \right]$, and the ECE by $\mathbb E_{\vec{x}} \left[ | \Delta_{\ferm(\vec x)}| \right]$.

Note that logistic regression can also be seen as the maximum likelihood estimator for the logit model, which given the data model \eqref{eq:def:data} for $\sigma_{\star}\neq \sigma$ is misspecified. \cite{Sur2019} have shown that even in the well-specified case $\sigma_{\star}=\sigma$, non-regularized logistic regression yields a biased estimator of $\vec{w}_{\star}$ in the high-dimensional limit where $n,d\to\infty$ at a proportional rate $\alpha=\sfrac{n}{d}$, which \cite{bai_dont_2021} has shown to be overconfident.  \cite{clarte_theoretical_2022} characterized the calibration as a function of the regularization strength and the number of samples, and has shown that overconfidence can be mitigated by properly regularizing. 

The goal in this section is to leverage these results on high-dimensional logistic regression in order to provide theoretical results on the calibration properties of TS and EC. In particular, we will be interested in comparing the following three choices of data likelihood function $\sigma_{\star}$:  
\begin{align}
\label{eq:def:activations}
    \begin{cases}
        \sigma_{\logit}(z) \!\!\!\!\!\!&= \frac{1}{1 + e^{-z}} \\
        \sigma_{\affine}(z) \!\!\!\!\!\!&= 0 \text{ if } z < - 1, 1 \text{ if } z > 1, \frac{t+1}{2} \text{ else }\\
        \sigma_{\constant} \!\!\!\!\!\! &= 0 \text{ if } z < -1, 1 \text{ if } z > 1, \sfrac{1}{2} \text{ else } \\
    \end{cases}
\end{align}

\begin{figure}[!t]
    \centering
    \def\figwidth{0.3\columnwidth}
    \def\figheight{0.3\columnwidth}

\begin{tikzpicture}
\tikzstyle{every node}=[font=\tiny]

\definecolor{color0}{rgb}{0.12156862745098,0.466666666666667,0.705882352941177}
\definecolor{color1}{rgb}{1,0.647058823529412,0}

\begin{axis}[
height=\figheight,
tick align=outside,
tick pos=left,
width=\figwidth,
x grid style={white!69.0196078431373!black},
xlabel={\(\displaystyle \alpha\)},
xmajorgrids,
xmin=0.0499999999999999, xmax=20.95,
xtick style={color=black},
y grid style={white!69.0196078431373!black},
ylabel={\(\displaystyle \sfrac{|T_{TS} - T_{EC}|}{T_{TS}} \)},
ymajorgrids,
ymin=0.00038490210650353, ymax=0.5,
ytick style={color=black}
]
\addplot [draw=black, fill=black, forget plot, mark=*, only marks, mark size=1pt]
table{%
x  y
1 0.000462023150714503
1.48717948717949 0.000602470093450275
1.97435897435897 0.000809277667252931
2.46153846153846 0.00121151856437121
2.94871794871795 0.00148509442251603
3.43589743589744 0.00150379077436698
3.92307692307692 0.00174345553337398
4.41025641025641 0.00190894518958171
4.8974358974359 0.0019427440471961
5.38461538461538 0.00179804565430677
5.87179487179487 0.00197873926395673
6.35897435897436 0.00198700708936501
6.84615384615385 0.00200444403493396
7.33333333333333 0.00199972430209456
7.82051282051282 0.00200000185184167
8.30769230769231 0.00190563835105904
8.7948717948718 0.0019023209792755
9.28205128205128 0.00195209581057124
9.76923076923077 0.00196308754279534
10.2564102564103 0.00184626217103798
10.7435897435897 0.00184756017439881
11.2307692307692 0.00182430003276102
11.7179487179487 0.00178382104814397
12.2051282051282 0.00183252553936068
12.6923076923077 0.00173862162413878
13.1794871794872 0.00188066094503442
13.6666666666667 0.00174355690968924
14.1538461538462 0.00168621316200239
14.6410256410256 0.00175553334967942
15.1282051282051 0.00171353098965743
15.6153846153846 0.00156028623576021
16.1025641025641 0.0017585991331515
16.5897435897436 0.00180168550720455
17.0769230769231 0.00151835487188877
17.5641025641026 0.00153122778474199
18.0512820512821 0.00167011001454102
18.5384615384615 0.00142231031017206
19.025641025641 0.00138845133092015
19.5128205128205 0.00150543908662474
20 0.00148839920131915
};
\addplot[draw=color0, fill=color0, forget plot, mark=*, only marks, mark size=1pt]
table {%
1 0.0121015384483389
1.48717948717949 0.0159600383446919
1.97435897435897 0.0191379697770742
2.46153846153846 0.0218273914421418
2.94871794871795 0.0242399284918305
3.43589743589744 0.0270653901277244
3.92307692307692 0.0337562329349698
4.41025641025641 0.0393238758215561
4.8974358974359 0.0438338267196012
5.38461538461538 0.0477204032964127
5.87179487179487 0.051165773594923
6.35897435897436 0.0542587352227779
6.84615384615385 0.0570852418020452
7.33333333333333 0.0596505475625118
7.82051282051282 0.0620363275491453
8.30769230769231 0.0642206437942322
8.7948717948718 0.0662326425029536
9.28205128205128 0.0681187046076588
9.76923076923077 0.0698773201033472
10.2564102564103 0.0715159347105361
10.7435897435897 0.0730407700015985
11.2307692307692 0.0744731146815771
11.7179487179487 0.0758332949223246
12.2051282051282 0.0770873412677037
12.6923076923077 0.0782941007393231
13.1794871794872 0.0794204349156341
13.6666666666667 0.0804866868669766
14.1538461538462 0.0814965093681872
14.6410256410256 0.0824545121858343
15.1282051282051 0.0833701093790081
15.6153846153846 0.0842398556085927
16.1025641025641 0.0850568541334762
16.5897435897436 0.0858451648127864
17.0769230769231 0.0865707636992444
17.5641025641026 0.0872894891576945
18.0512820512821 0.0879731410865527
18.5384615384615 0.0886248284415442
19.025641025641 0.089237879519549
19.5128205128205 0.0898340761906901
20 0.0904058617632092
};
\addplot[draw=color1, fill=color1, forget plot, mark=*, only marks, mark size=1pt]
table{%
x  y
1 0.0156359335601499
1.48717948717949 0.0125569727997651
1.97435897435897 0.024722736211889
2.46153846153846 0.0270668866741009
2.94871794871795 0.0403460127481245
3.43589743589744 0.0341715704259338
3.92307692307692 0.0766364811595017
4.41025641025641 0.088671818131378
4.8974358974359 0.119626485217223
5.38461538461538 0.10012117111977
5.87179487179487 0.128211157409077
6.35897435897436 0.106384226513938
6.84615384615385 0.126680284698584
7.33333333333333 0.130126344766854
7.82051282051282 0.138637360633119
8.30769230769231 0.140640026493109
8.7948717948718 0.171459662755168
9.28205128205128 0.158991963470476
9.76923076923077 0.180785312127454
10.2564102564103 0.172381646859328
10.7435897435897 0.168272990915476
11.2307692307692 0.192660002325984
11.7179487179487 0.192450266435639
12.2051282051282 0.201839230626814
12.6923076923077 0.211749005300134
13.1794871794872 0.206411711595742
13.6666666666667 0.215771394740272
14.1538461538462 0.195467857424946
14.6410256410256 0.231796320874372
15.1282051282051 0.236815406251558
15.6153846153846 0.246719260711447
16.1025641025641 0.256777404911938
16.5897435897436 0.245652444085573
17.0769230769231 0.224028783229418
17.5641025641026 0.288434989860511
18.0512820512821 0.261818118713328
18.5384615384615 0.266998126403211
19.025641025641 0.289696478084658
19.5128205128205 0.266643089629938
20 0.258647666721105
};
\addplot [semithick, black]
table {%
1 0.000589988634129368
1.48717948717949 0.000708293705587901
1.97435897435897 0.000789964136126098
2.46153846153846 0.001098228616311
2.94871794871795 0.00140589933023946
3.43589743589744 0.00159943873245626
3.92307692307692 0.00173668273838204
4.41025641025641 0.00182986815393864
4.8974358974359 0.00189443987386998
5.38461538461538 0.00193775950681515
5.87179487179487 0.00197484306400807
6.35897435897436 0.00198390556938507
6.84615384615385 0.00200247613826795
7.33333333333333 0.00199956410536413
7.82051282051282 0.00199476719269664
8.30769230769231 0.00198560088678506
8.7948717948718 0.00197806079720671
9.28205128205128 0.00195526915171475
9.76923076923077 0.00193480348411761
10.2564102564103 0.00191223752558322
10.7435897435897 0.0018918638811322
11.2307692307692 0.00187167064545975
11.7179487179487 0.00184795191882112
12.2051282051282 0.00182888223736288
12.6923076923077 0.00180047809070184
13.1794871794872 0.00177348564779448
13.6666666666667 0.00176897851038554
14.1538461538462 0.00174443552656251
14.6410256410256 0.0017202057058638
15.1282051282051 0.00169622046516174
15.6153846153846 0.00167254566766779
16.1025641025641 0.00164857249604391
16.5897435897436 0.00162405692698497
17.0769230769231 0.00160299566577993
17.5641025641026 0.00158723470411071
18.0512820512821 0.00155571386972682
18.5384615384615 0.00154709710135647
19.025641025641 0.0015077449621815
19.5128205128205 0.00149978722023299
20 0.00148024056246153
};
\addlegendentry{$\sigma_{logit}$}
\addplot [semithick, color0]
table {%
1 0.0121015384483389
1.48717948717949 0.0159600383446919
1.97435897435897 0.0191379697770742
2.46153846153846 0.0218273914421418
2.94871794871795 0.0242399284918305
3.43589743589744 0.0270653901277244
3.92307692307692 0.0337562329349698
4.41025641025641 0.0393238758215561
4.8974358974359 0.0438338267196012
5.38461538461538 0.0477204032964127
5.87179487179487 0.051165773594923
6.35897435897436 0.0542587352227779
6.84615384615385 0.0570852418020452
7.33333333333333 0.0596505475625118
7.82051282051282 0.0620363275491453
8.30769230769231 0.0642206437942322
8.7948717948718 0.0662326425029536
9.28205128205128 0.0681187046076588
9.76923076923077 0.0698773201033472
10.2564102564103 0.0715159347105361
10.7435897435897 0.0730407700015985
11.2307692307692 0.0744731146815771
11.7179487179487 0.0758332949223246
12.2051282051282 0.0770873412677037
12.6923076923077 0.0782941007393231
13.1794871794872 0.0794204349156341
13.6666666666667 0.0804866868669766
14.1538461538462 0.0814965093681872
14.6410256410256 0.0824545121858343
15.1282051282051 0.0833701093790081
15.6153846153846 0.0842398556085927
16.1025641025641 0.0850568541334762
16.5897435897436 0.0858451648127864
17.0769230769231 0.0865707636992444
17.5641025641026 0.0872894891576945
18.0512820512821 0.0879731410865527
18.5384615384615 0.0886248284415442
19.025641025641 0.089237879519549
19.5128205128205 0.0898340761906901
20 0.0904058617632092
};
\addlegendentry{$\sigma_{affine}$}
\addplot [semithick, color1]
table {%
1 0.0127140893391908
1.48717948717949 0.0167927874416053
1.97435897435897 0.0203033734226291
2.46153846153846 0.027070757971925
2.94871794871795 0.0429792801127694
3.43589743589744 0.0557869808616585
3.92307692307692 0.0675170526820274
4.41025641025641 0.0785768206936381
4.8974358974359 0.0891212245840087
5.38461538461538 0.099198969861594
5.87179487179487 0.108854850928242
6.35897435897436 0.118108934352487
6.84615384615385 0.127010632678464
7.33333333333333 0.135536364731892
7.82051282051282 0.143738994055264
8.30769230769231 0.151607764854664
8.7948717948718 0.15916601936948
9.28205128205128 0.166452129517935
9.76923076923077 0.173435209952785
10.2564102564103 0.180147663568115
10.7435897435897 0.186599686689499
11.2307692307692 0.192812702474115
11.7179487179487 0.198769910141708
12.2051282051282 0.204510831406729
12.6923076923077 0.210050533888488
13.1794871794872 0.215371228032846
13.6666666666667 0.220494103120885
14.1538461538462 0.225437667321066
14.6410256410256 0.230179144935238
15.1282051282051 0.234758825369641
15.6153846153846 0.239169712926581
16.1025641025641 0.243424700395972
16.5897435897436 0.247529763623118
17.0769230769231 0.251492039611132
17.5641025641026 0.255319182320396
18.0512820512821 0.259018099119162
18.5384615384615 0.262570394203155
19.025641025641 0.26601579791989
19.5128205128205 0.269344895606136
20 0.272579607023013
};
\addlegendentry{$\sigma_{constant}$}

\end{axis}

\end{tikzpicture}
    \caption{Relative difference $\sfrac{| T_{EC} - T_{TS} |}{T_{TS}}$ as a function of the sampling ratio $\alpha$ with three different $\sigma_{\star}$, and $\lambda = 10^{-4}$. We observe that when $\sigma_{\star}$ differs more from $\sigma$, EC and TS yield different results. Points are simulations done at $d = 200$. }
    \label{fig:relative_difference}
\end{figure}
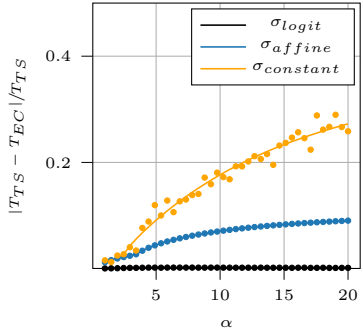

\paragraph{Asymptotic uncertainty metrics ---} The starting point of the analysis is to note that the uncertainty metrics of interest \eqref{eq:def:metrics} only depend on the weights through the pre-activations $(\vec{w}_{\star}^{\top}\vec{x}, \hat{\vec{w}}^{\top}\vec{x})$ on a test point $\vec{x}$. Since the distribution of the inputs is Gaussian, the joint statistics of the pre-activations is Gaussian:
\begin{align*}
(\vec{w}_{\star}^{\top}\vec{x}, \hat{\vec{w}}^{\top}\vec{x}) \sim \mathcal{N}\left(\vec{0}_{2},
\begin{bmatrix}
    \sfrac{1}{d}||\vec{w}_{\star}||^{2}_{2} & \sfrac{1}{d}\vec{w}_{\star}^{\top}\hat{\vec{w}}_{\erm}\\ \sfrac{1}{d}\vec{w}_{\star}^{\top}\hat{\vec{w}}_{\erm} & \sfrac{1}{d}||\hat{\vec{w}}_{\erm}||^{2}_{2}
\end{bmatrix}
\right)
\end{align*}

\begin{figure*}[!ht]
    \centering
    \def\figwidth{0.5\columnwidth} 
    \def\figheight{0.5\columnwidth}

    \includegraphics[height=0.3\columnwidth]{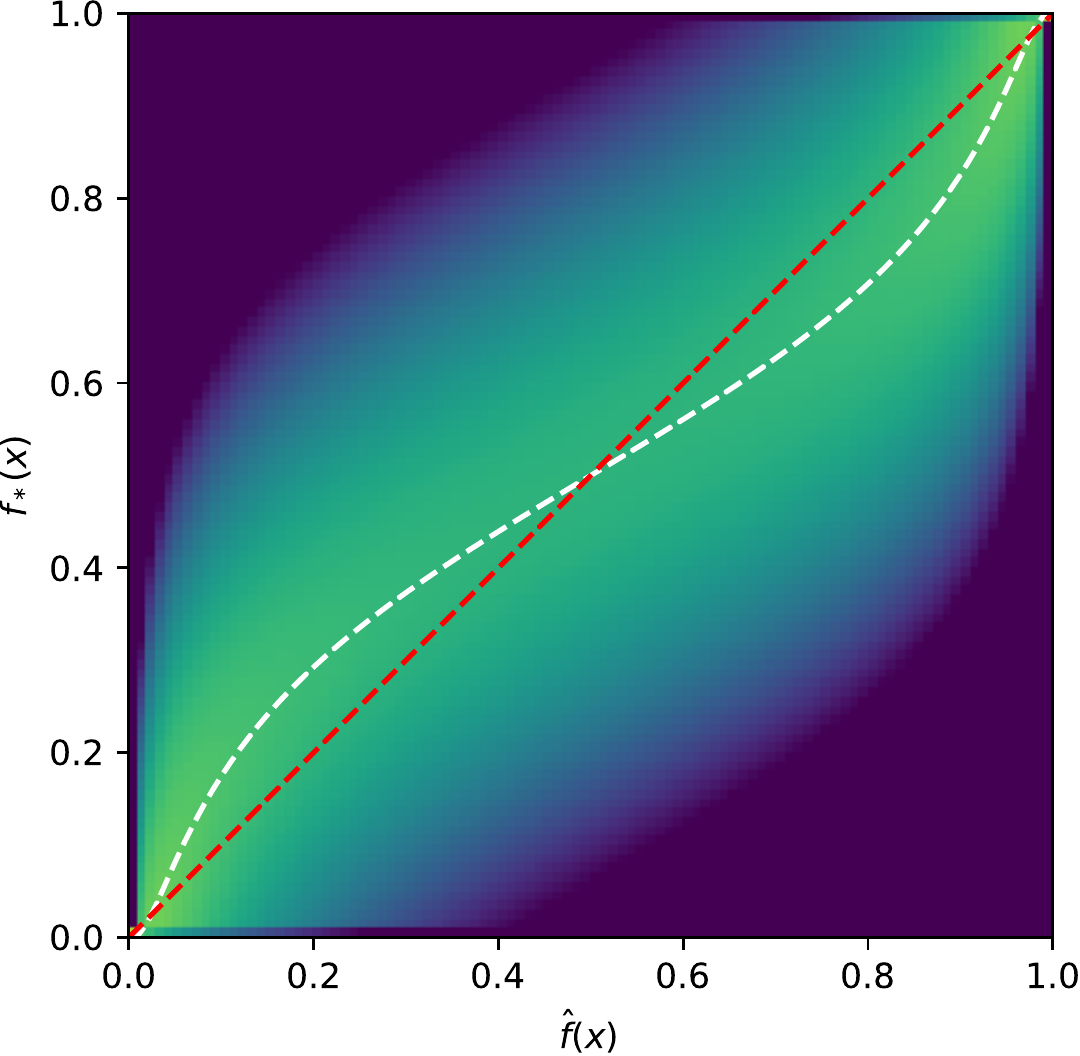}
    \includegraphics[height=0.3\columnwidth]{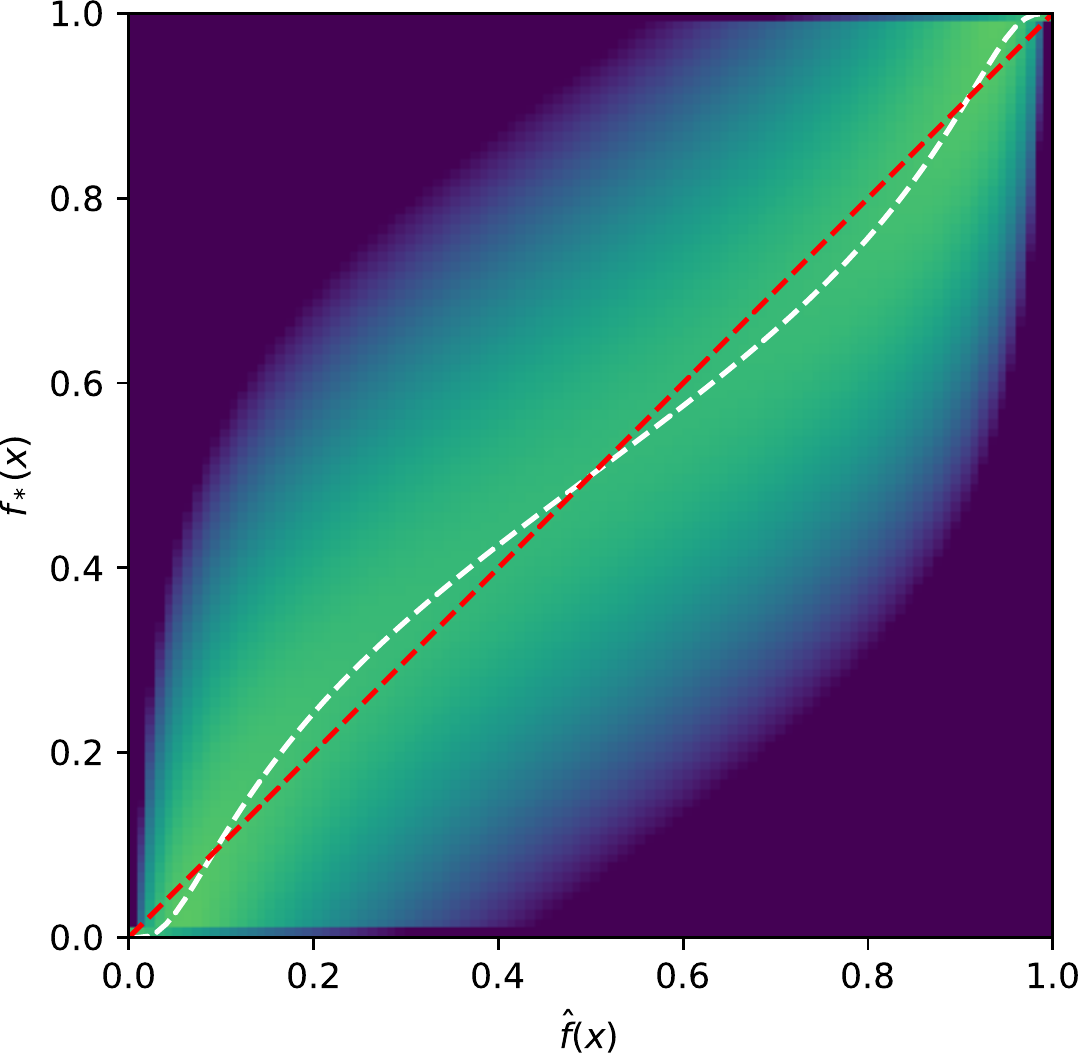}
    \includegraphics[height=0.3\columnwidth]{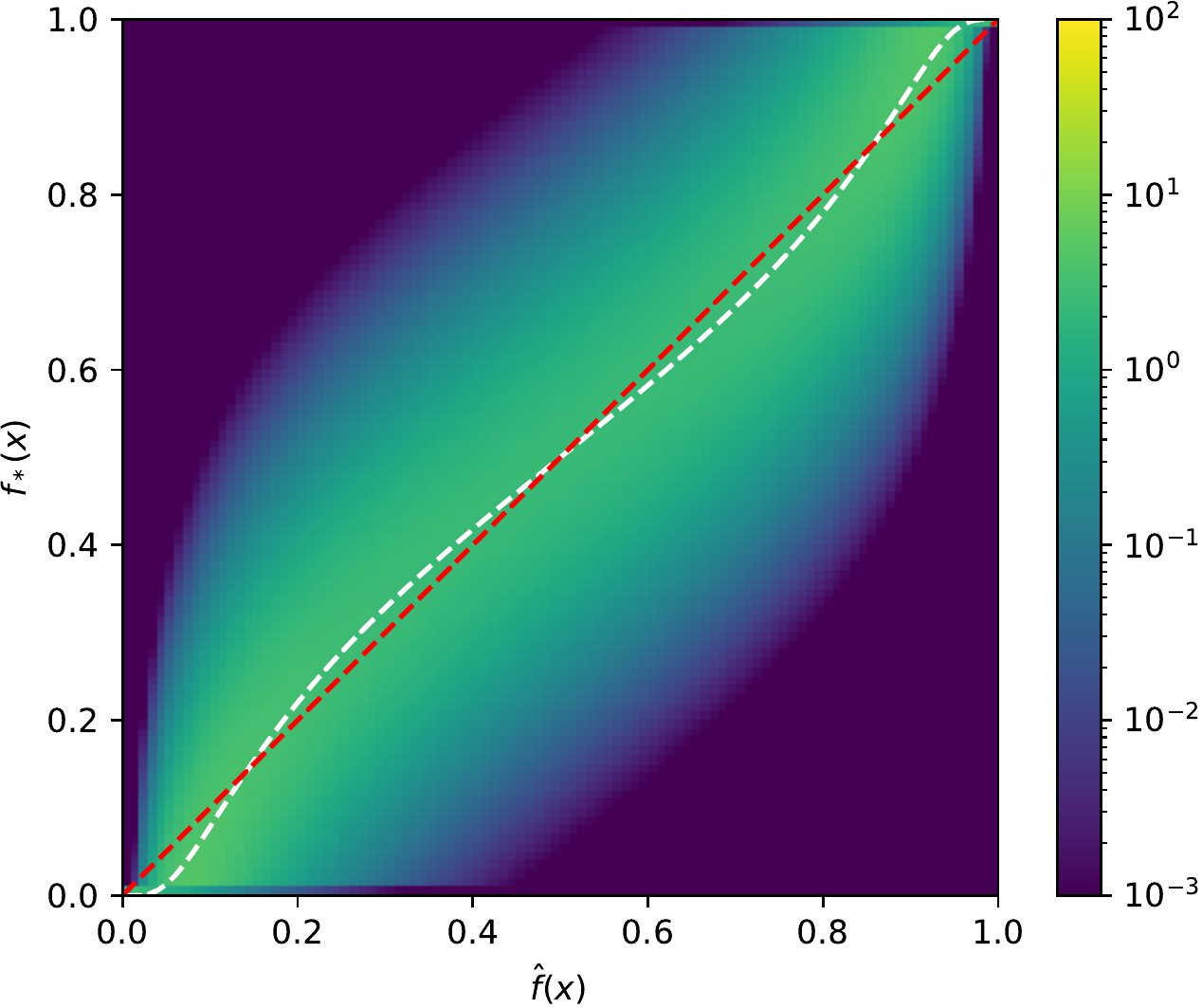}
    \caption{Plots of the density of $(\ferm(\vec{x}), f_{\star}(\vec{x}))$ (Left), after Temperature scaling (Middle) and expectation consistency (Right), for the sampling ratio $\sfrac{n}{d} = 20$ and regularization $\lambda = 10^{-4}$. Dashed white lines represent the accuracy as a function of the confidence, the red line is the diagonal. The difference between red and white lines corresponds to the calibration. ECE of $\ferm$ is, from left to right: 2.1 \%, 1.2 \%, 1.0 \%. We have $T_{TS} = 1.24, T_{EC} = 1.35$. }
    \label{fig:density_plots} 
\end{figure*}

As discussed above, different recent works \citep{Sur2019, bai_dont_2021, clarte_theoretical_2022} have derived exact asymptotic formulas for these statistics in different levels of generality for logistic regression. In particular, the following theorem from \cite{clarte_theoretical_2022}, which considers a general misspecified model will be used for the analysis:  
\begin{theorem}[Thm. 3.2 from \cite{clarte_theoretical_2022}] 
\label{thm:stats}
Consider the logit classifier \eqref{eq:def:logit} trained by minimizing the empirical risk \eqref{eq:def:risk} on a data set $(\vec{x}_{i},y_{i})_{i\in[n]}$ independently sampled from model \eqref{eq:def:data}. Then, in the high-dimensional limit when $n,d\to\infty$ at fixed $\alpha=\sfrac{n}{d}$:
\begin{align}
\label{eq:def:overlaps}
    (\sfrac{1}{d}\vec{w}_{\star}^{\top}\hat{\vec{w}}_{\erm}, \sfrac{1}{d}||\hat{\vec{w}}_{\erm}||^{2}_{2}) \xrightarrow[d\to\infty]{} (m,q)
\end{align}
where $(m,q)\in\mathbb{R}_{+}^{2}$ are explicitly given by the solution of a set of low-dimensional self-consistent equations depending only on $(\alpha, \lambda, \sigma,\sigma_{*})$, and which for the sake of space are discussed in Appendix B.
\end{theorem}
Leveraging on Thm. \ref{thm:stats}, we can derive an asymptotic characterization for the asymptotic limit of the uncertainty metrics defined in \eqref{eq:def:metrics}.
\begin{proposition}
\label{thm:metrics}
Under the same assumptions of Theorem \ref{thm:stats}, the asymptotic limit of the uncertainty metrics defined in \eqref{eq:def:metrics} is given by:
\begin{align*}
\begin{cases}
    \Delta_{\ell}(m, q) &= \ell - \mathcal{Z}_{\star}(1, \sfrac{m}{q}\sigma^{-1}(\ell), 1 - \sfrac{m^2}{q}) \\
    {\rm ECE}(m, q) &= \int_{0}^{\infty} \dd z | \Delta_{\sigma(z)}(m, q) | \mathcal{N}(z | 0, q) \\
\end{cases}
    \label{eq:state_evolution_bs}
\end{align*}
where $(m, q)\in\mathbb{R}^{2}_{+}$ are the asymptotic limits of the correlation functions in \eqref{eq:def:overlaps} and
\begin{align}
    \mathcal{Z}_{\star}(y, \omega, V) = \mathbb{E}_{\xi\sim\mathcal{N}(\omega, V)}\left[\sigma_{\star}\left(\sfrac{y\xi}{T_{\star}}\right)\right]
\end{align}
\end{proposition}
The proof of this result is given in Appendix~B. Proposition \ref{thm:metrics} provides us with all we need to fully characterize the calibration properties of TS and EC in our setting. In the next paragraphs, we discuss its implications.

In practice, the $\ell_2$ regularization parameter $\lambda$ in the empirical risk \eqref{eq:def:risk} is optimized by cross-validation. \cite{clarte_theoretical_2022, clarte_study_2022} has shown that appropriately regularizing the risk not only improves the prediction accuracy but also the calibration and ECE of the logistic classifier. In particular, it was shown that cross-validating on the loss function yields different results from cross-validation on the misclassification error, with a larger difference arising in the case of misspecified models. Curiously, \cite{clarte_study_2022} has shown that in this case, good performance and calibration can be achieved by combining a $\ell_{2}$ penalty with TS. In the following, we discuss how this compares with EC. Note that the exact asymptotic characterization from Thm.~\ref{thm:stats} allows us to bypass cross-validation, allowing us to find the optimal $\lambda$ by directly optimizing the low-dimensional formulas. We thus define $\lambdaerror$ (respectively $\lambdaloss$) as the value of $\lambda$ such that $\werm$ yields the lowest test misclassification error (respectively test loss).

\subsection{EC outperforms TS} 
\label{sec:theory:outperforms}
In Section~\ref{sec:real_data}, we have numerically observed that EC and TS yield almost the same temperature and thus have similar performance in terms of different uncertainty quantification metrics for different architectures trained on real data sets. Figure~\ref{fig:relative_difference} shows the relative difference $\delta T = \sfrac{|T_{TS} - T_{EC}|}{T_{TS}}$ between the two methods for logistic regression on the synthetic data model \eqref{eq:def:data} for the different choice of target activation $\sigma_{\star}\in\{\sigma_{\logit}, \sigma_{\affine}, \sigma_{\constant}\}$ defined in \eqref{eq:def:activations}. Contrary to the real data scenario in Section~\ref{sec:real_data}, we observe a significant difference between the two methods for $\sigma_{\star}\in\{\sigma_{\affine}, \sigma_{\constant}\}$. For instance, for the piece-wise constant function $\sigma_{\star} = \sigma_{\constant}$, $\delta T$ is a non-decreasing function of the sampling ratio $\alpha$, and is around $30 \%$ at $\alpha = 20$.

Figure~\ref{fig:ece} shows that expectation consistency yields a lower ECE than Temperature scaling in all the settings considered in Section \ref{sec:theory_method}. On one hand, the effect is small in the well-specified case where the target and model likelihoods are the same: the ECE of Temperature scaling is higher by around $0.01\%$. This is quite intuitive from the discussion in Section \ref{sec:bayesian}, since in this case, we are closer to the Bayesian setting where both methods were shown to coincide. On the other hand, this difference increases in the misspecified setting, suggesting that model misspecification plays an important role in these calibration methods. In particular, note that in all cases considered here, EC has a lower ECE than TS for all three regularizations considered: $\lambda = 10^{-4}, \lambdaerror, \lambdaloss$.

Figure~\ref{fig:density_plots} shows the joint probability density function of the variables $(f_{\star}(\vec{x}), \ferm(\vec{x})) \in [0, 1]^2$. In particular, we show in white-dashed lines the conditional mean $\mathbb{E} \left[ \fstar(\vec{x}) | \ferm(\Vec{x}) \right]$ which corresponds to the accuracy-confidence chart in Figure~\ref{fig:real_data_curves}. As in the real data case, we observe that the ERM estimator is consistently overconfident, i.e $\forall \ell \geqslant \sfrac{1}{2},  \Delta_{\ell} \geqslant 0$. Moreover, we see that after TS and EC, the conditional mean gets closer to the diagonal (red curve), implying that the model is more calibrated. The phenomenology of the simple data model seems to correspond to what we observe with real data and suggests that expectation consistency is a better approach to calibration.

\paragraph{Interpretation of the results ---} Temperature scaling corresponds to rescaling the outputs of the network by minimizing the validation loss. In the literature, the cross-entropy loss is one of the most widespread choices, both for training and for measuring uncertainty scores (with the softmax). From a Bayesian perspective, minimizing the cross-entropy loss corresponds to maximizing the likelihood under the assumption that the has been generated from a softmax (a.k.a. multinomial logit) model. Hence, the underlying assumption behind temperature scaling is that the labels are generated using a softmax likelihood. Therefore, we expect it to perform better when this assumption is met. Indeed, our experiments in Section \ref{sec:theory_method} confirm this intuition. In the case where the ground truth model is indeed given by a logit, TS performs well and is close to EC. However, in the misspecified case, where this assumption does not hold, TS performs worse than EC.

\section{Conclusion and future work} 
\label{sec:conclusion}
In this work, we introduced \textit{Expectation Consistency}, a new post-training calibration method  for neural networks. We have shown that EC is close to temperature scaling across different image classification tasks, giving almost the same expected calibration error and Brier score, while having comparable computational cost. Additionally, we provided an analysis of the asymptotic properties of both methods in a synthetic setting where data is generated by a ground truth model, showing that while EC and TS yield the same performance for well-specified methods, EC provides a better and more principled calibration method under model misspecification.

Our experiments on simple data models showed that when there is a discrepancy between our linear model and the true data model, EC performs better than TS. However, our experiments on real data show a very similar performance across different architectures, data models and overall model accuracy. In future work we aim to understand better why both methods are so similar in practical scenarios.

\section{Acknowledgements} 
\label{sec:acknowledgements}
We acknowledge funding from the ERC under the European Union’s Horizon 2020 Research and Innovation Program Grant Agreement 714608-SMiLe, the Swiss National Science Foundation grant SNFS OperaGOST, $200021\_200390$ and the \textit{Choose France - CNRS AI Rising Talents} program. This research was supported by the NCCR MARVEL, a National Centre of Competence in Research, funded by the Swiss National Science Foundation (grant number 205602).
\clearpage

\bibliography{refs}

\clearpage
\appendix

\section{Details on training procedure}
\label{sec:training_procedure}

\paragraph{SVHN} For the SVHN dataset \cite{netzer_reading_2011}, the Resnet20 model of depth 20 and containing 0.27M parameters was trained for $50$ epochs, using SGD with a learning rate $\eta = 0.1$, weight decay $1e-4$ and momentum $0.9$. $90\%$ of data points were used for training and the rest was used for validation.
\paragraph{CIFAR10} ResNet models (of depth 20, 56 with Resnet56 having 0.85M parameters) were trained for $50$ epochs, using SGD with a learning rate $\eta = 0.1$, weight decay $1e-4$ and momentum $0.9$. The DenseNet 121 (containing 7.9 parameters) was trained with the same parameters as the ResNets, except for the learning rate $\eta = 0.01$. As in \cite{he_resnet_2016}, images in the training set were randomly cropped and flipped horizontally.

\paragraph{CIFAR100} On CIFAR100, we used pre-trained models from the Github repository \url{https://github.com/chenyaofo/pytorch-cifar-models}. These models were trained on the entirety of the training set, so the test set containing $10000$ images was split in half into a validation and test set, containing $5000$ images each.

\subsection{Additional plots}

\begin{figure}[h!]
    \centering
    \includegraphics[width=0.45\textwidth]{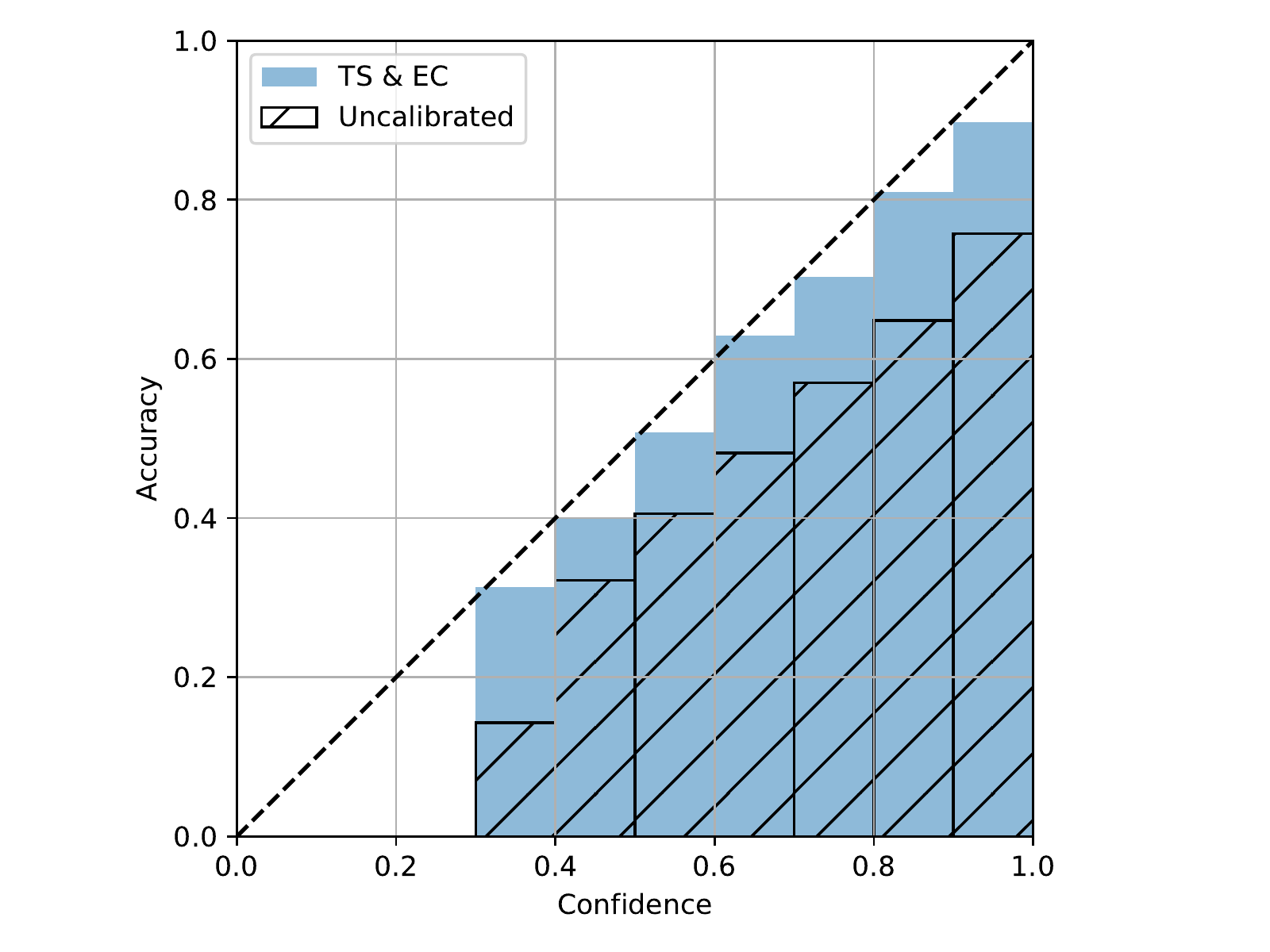}
    \includegraphics[width=0.45\textwidth]{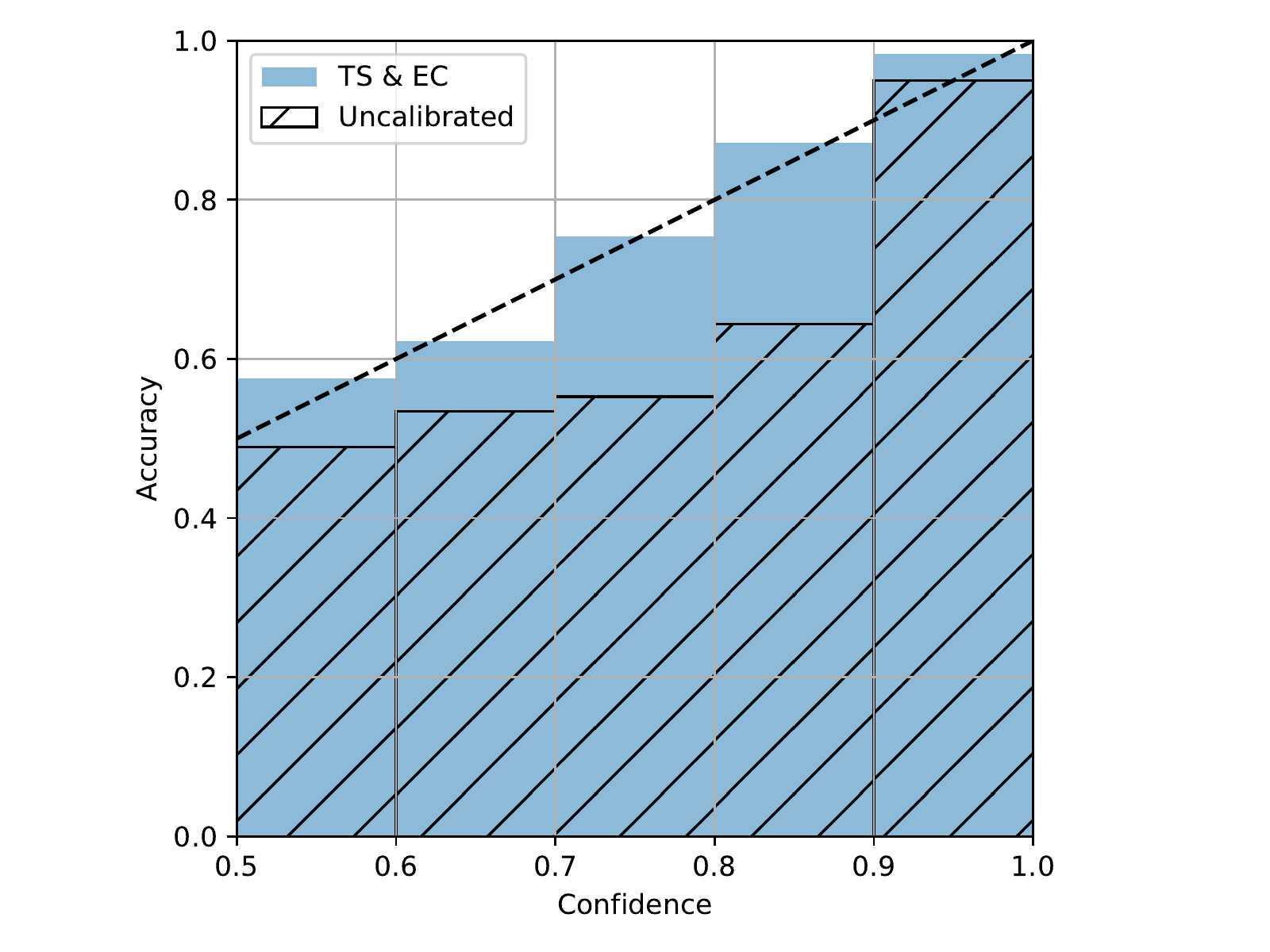}
    \caption{Reliability diagram of Resnet20 on the SVHN dataset (Left) and Resnet56 on the CIFAR10 dataset (Right). Before calibration, both methods are overconfident. TS and EC improve calibration and mitigate overconfidence. }
    \label{app:reliability_diagram} 
\end{figure}

In Figure~\ref{app:reliability_diagram}, we plot the reliability diagram of Resnet20 and Resnet56 on SVHN and CIFAR10 respectively. We observe that the uncalibrated models are overconfident (as the confidence is higher than the corresponding accuracy), and both TS and EC mitigate this overconfidence.

\section{State evolution equation}
\label{app:se_equations}

In this section, we focus on the data model introduced in Section 5. Recall that we consider a dataset of $n$ samples $\mathcal{D} = (x^{\mu}, y^{\mu})_{\mu = 1}^n$ generated by 
\begin{equation}
    \Vec{x} \sim \mathcal{N}(\Vec{0}, \mathcal{I}_d / d), \Vec{w}_* \sim \mathcal{N}(\Vec{0}, \mathcal{I}_d), \mathbb{P}(y = 1 | \wstar^{\top} \Vec{x}) = \sigma_{\star}(\wstar^{\top} \Vec{x})
\end{equation}
and we fit the following logistic regression model, with $\sigma$ the sigmoid function: 
\begin{equation}
    \ferm(\Vec{x}) = \sigma(\werm^{\top} \Vec{x})
\end{equation}
by minimizing the following empirical risk 
\begin{equation}
    \mathcal{R}(\Vec{w}, \mathcal{D}, \lambda) = \sum_{\mu = 1}^n \log \sigma(y^{\mu} \Vec{w}^{\top} \Vec{x}) + \sfrac{\lambda}{2} \| \Vec{w} \|^2
\end{equation}
we thus have $\werm = \arg \min_{\Vec{w}} \mathcal{R}(\Vec{w}, \mathcal{D}, \lambda)$. For a new sample $\Vec{x}$, we are interested in the joint distribution of $\fstar(\Vec{x})$ and $\ferm(\Vec{x})$. As these two functions only depend on the scalar products $\wstar^{\top}\Vec{x}$, $\werm^{\top}\Vec{x}$ it suffices to compute the joint distribution of these scalar products. By the Gaussianity of $\Vec{x}$, we just need to compute the \textit{overlaps} $m = \wstar^{\top} \werm$ and $q = \| \werm \|^2$. In the high-dimensional limit where $n, d \to \infty$ but where we keep the \textit{sampling ratio} constant $\sfrac{n}{d} = \alpha$, it is possible to compute the value of $m$ and $q$. The idea is to introduce the distribution
\begin{equation}
    \mu_{\beta, \mathcal{D}, \lambda}(\Vec{w}) = \frac{1}{\mathcal{Z}_{\beta}} \exp \left( - \beta \mathcal{R}(\Vec{w}, \mathcal{D}, \lambda) \right)
\end{equation}
where $\mathcal{Z}_{\beta}$ is a normalization constant. In the limit $\beta \to \infty$, $\mu_{\beta, \mathcal{D}, \lambda}$ converges to a Dirac distribution peaked at $\werm = \arg \min \mathcal{R}(\Vec{w}, \mathcal{D}, \lambda)$. To compute $m, q$, one needs to compute the expression of $\log \mathcal{Z}_{\beta}$ and its limit when $\beta \to \infty$. In the high-dimensional regime where both the dimension and number of samples diverge with a fixed ratio, this can be done using the \textit{replica method} from statistical physics \cite{zdeborova2016statistical}. As these computations are not the focus of the present paper, we refer to \cite{loureiro_learning_2021, clarte_theoretical_2022} for the detailed computations. In the end, if we define 
\begin{align}
    \mathcal{Z_*}(y, \omega, v_*) &= \int \dd z \sigma_*(y \times z) \mathcal{N}( z | \omega, v_*) \\
    f(y, \omega, v) &= \arg \min_z \left[  \frac{(z - \omega)^2}{2v} - \log \sigma(z) \right]
\end{align}
then $m, q$ are the solution of the following self-consistent equations: 
\begin{align}
    \begin{cases}
        m &= \frac{\hat{m}}{\lambda + \hat{v}} \\
        q &= \frac{ \hat{q} + \hat{m}^2 }{(\lambda + \hat{v})^2}   \\
        v &= \frac{1}{\lambda + \hat{v}}
    \end{cases}, \quad 
    \begin{cases}
        \hat{m} &= \alpha \mathbb{E}_{ \xi \sim \mathcal{N}(0, q)} \left[ \int \dd y \partial_{\omega} \mathcal{Z}_*(y, \sfrac{m}{q}\xi, v_*) f(y, \xi, v) \right]  \\
        \hat{q} &= \alpha \mathbb{E}_{ \xi \sim \mathcal{N}(0, q) } \left[ \int \dd y \mathcal{Z}_*(y, \sfrac{m}{q}\xi, v_*) f^2(y, \xi, v) \right] \\
        \hat{v} &= - \alpha \mathbb{E}_{ \xi \sim \mathcal{N}(0, q) } \left[ \int \dd y \mathcal{Z}_*(y, \sfrac{m}{q}\xi, v_*) \partial_{\omega} f(y, \xi, v) \right]
    \end{cases}
\end{align}
with $v_* = \rho - \sfrac{m^2}{q}$.

\paragraph{Calibration in the high-dimensional regime} Once we obtained the overlaps $m, q$, we can derive the expression the calibration $\Delta_{\ell}$: 
\begin{align}
    \Delta_{\ell} &= \mathbb E \left[ \fstar(\vec x) | \ferm(\vec x) \right] = \mathbb P \left[ y = 1 | \ferm(\vec x) \right] = \int \dd z \sigma_{\star}(z) \mathcal{N}(z | \frac{m}{q} \ferm^{-1}(\vec{x}), \rho - \sfrac{m^2}{q})
\end{align} 
The second line comes from the fact that the scalar product $\wstar^{\top} \Vec{x}$ conditioned on $\werm^{\top} \Vec{x} = \sigma^{-1}(\ell)$ follows a Gaussian distribution with mean $\sfrac{m}{q} \xi$ and variance $\rho - \sfrac{m^2}{q}$. As a consequence, the expression of ECE is 
\begin{align}
    ECE &= \mathbb{E}_{\Vec{x}} \left[ | \Delta_{\ferm(\Vec{x})}| \right] = \mathbb{E}_{\xi = \werm^{\top}\Vec{x}} \left[ | \Delta_{\sigma(\xi)}|\right] = \int \dd \xi |\Delta_{\sigma(\xi)}| \mathcal{N}(\xi | 0, q)
\end{align}

\section{Experiments on corrupted dataset}

We describe below an experiment where EC can significantly improve over TS for real data: we train different architectures on several image classification tasks, as in Figure 1. However, here for the validation and test set some classes are replaced with random labels. For SVHN and CIFAR10, the labels $y = 0$ are replaced by random labels. For CIFAR100, the labels $y = 0, ..., 9$ are replaced by random labels. By doing so, around 10\%  of validation/test data is corrupted, with a noise that depends on the class. 
Note that the training data is left unchanged: the goal of this experiment is to model a distribution shift between training and test data, similarly as what is done \cite{hendrycks_benchmarking_2019}.

In the table below, we compare the performance (in ECE and Brier score) of EC and TS with these corrupted datasets. We observe that in this setting, EC outperforms TS by a significant margin on several datasets and architectures.

\begin{table*}
    \centering
    \begin{tabular}{c|c|ccc|ccc|ccc}
        \toprule
        Dataset & Model & $\mathcal{E}_g$ & $T_{TS}$ & $T_{EC}$ & $ECE$ & \!\!\!\! $ECE_{TS}$ \!\!\!\!& \!\!\!\!$ECE_{EC}$ \!\!\!\!& \!\!\!\! $BS$ \!\!\!\! & \!\!\!\!$BS_{TS}$ \!\!\!\!& \!\!\!\!$BS_{EC}$ \!\!\!\!\\
        \midrule
        SVHN & Resnet20 & 12.5 \%  & 2.69 & 2.23 & 8.3 \% & 10.7 \% & 7.5 \% & 21.9 \% & 23.4 \%  &  22.1 \%  \\ 
        \midrule
        CIFAR10  & Resnet20 & 20.9 \% & 2.4 & 2.0 & 12.8 \% & 4.6 \% & 4.2 \% & 34.2 \% & 32.2 \% & 32.1 \%\\
        CIFAR10  & Resnet56 &  21 \% & 2.58 & 2.15 & 13.8 \% & 5.4 \% & 4.9 \%& 35.2 \% & 32.9 \% & 32.8 \% \\
        CIFAR10  & Densenet121 & 20.4 \% & 2.76 & 2.54 & 15.8 \% & 3.6 \%& 5.0 \%& 35.9 \% & 31.8 \% & 31.9 \%\\
        \midrule
        CIFAR100 & Resnet20 & 38.1 \% & 2.04 & 1.70 & 16.5 \%& 9.6 \% & 5.9 \% & 57.0 \% & 54.9 \% & 53.9 \% \\
        CIFAR100 & Resnet56 & 34.8 \% & 2.27 & 2.10 & 21.7 \%& 7.6 \% & 7.3 \% & 56.0 \% & 50.6 \% & 50.4 \%  \\ 
        CIFAR100 & VGG19 & 35.5 \% & 2.6 & 2.1 & 28.34 \% & 5.2   & 5.1 \% & 61.8 \% & 50.1 \% & 50.1 \% \\
        CIFAR100 & RepVGG-A2 & 30.5 \% & 1.44 & 1.40 & 13.7 \% & 11.6 \% & 11.7 \% & 47.2 \% &  47.1 \% & 47.0 \% \\
        \bottomrule
    \end{tabular}
    \caption{Comparison of expected calibration error (ECE) and Brier score (BS) of temperature scaling (TS) and expectation consistency (EC) when part of the validation and test data has been corrupted}
    \label{fig:tab_ece}
\end{table*}

\end{document}